\documentclass[runningheads]{llncs}

\usepackage{graphicx}
\usepackage{comment}
\usepackage{amsmath,amssymb}
\usepackage{color}
\usepackage{url}
\usepackage{hyperref}
\usepackage{array}
\usepackage{tabularx}
\usepackage{lipsum}
\usepackage{graphicx}
\usepackage{subcaption}
\usepackage{hhline}
\usepackage{tikz}
\usepackage{stfloats}
\usepackage{multirow}
\usepackage{pgfplots}
\pgfplotsset{width=5cm,compat=1.9}
\usetikzlibrary{patterns}
 \usepackage[capitalise]{cleveref}

\usepackage[export]{adjustbox}

\usepackage{graphicx}
\usepackage{caption}
\usepackage{subcaption}
\usepackage{tikz}
\usetikzlibrary{external}
\usepackage{pgfplots}
\usetikzlibrary{spy,calc}
\usepackage{hyperref}

\newif\ifreview
\reviewfalse

\ifreview
	\usepackage{lineno}

	\linenumbers
\fi

\begin{document}

%%%%%%%%%%%%%%%%%%%%% Add submission id, track, and title. %%%%%%%%%%%%%%%%%%%%%

% TODO: Please insert your submission number here
\def\SubNumber{69}

% TODO: Please uncomment the track this paper will be submitted to, comment all other lines
%\def\GCPRTrack{Main Track}
%\def\GCPRTrack{Special Track: Pattern recognition in the life and natural sciences}
%\def\GCPRTrack{Special Track: Photogrammetry and remote sensing}
%\def\GCPRTrack{Young Researcher's Forum}
\def\GCPRTrack{Fast Review Track}

% TODO: Replace with your title
\title{Robust 3D Gaussian Splatting for Novel View Synthesis in Presence of Distractors}
% You can use \thanks for acknowledgment. Do not add any acknowledgment to the draft 
% version that is used for the review process.  
%\title{Title\thanks{XXX}}

\ifreview
	% ANONYMOUS SUBMISSION FOR REVIEW
	% DO NOT MODIFY these for the draft version that is used for the review process.
	\titlerunning{GCPR 2024 Submission \SubNumber{}. CONFIDENTIAL REVIEW COPY.}
	\authorrunning{GCPR 2024 Submission \SubNumber{}. CONFIDENTIAL REVIEW COPY.}
	\author{GCPR 2024 - \GCPRTrack{}}
	\institute{Paper ID \SubNumber}
\else
	\titlerunning{Robust 3D Gaussian Splatting}

	\author{Paul Ungermann \and
	Armin Ettenhofer\and
	Matthias Nie\ss{}ner \and
        Barbara Roessle}
	
	\authorrunning{P. Ungermann et al.}
	
	\institute{
 Technical University of Munich, Munich, Germany
	\email{\{paul.ungermann,armin.ettenhofer,niessner,barbara.roessle\}@tum.de}
 }
\fi

\maketitle              % typeset the header of the contribution

\vspace{-0.4cm}
\begin{figure*}[!h]
    \centering
        \centering
        \begin{subfigure}[t]{0.08\textwidth}
        \centering
        \rotatebox{90}{\parbox{1.6cm}{\centering Training \\ Images}}\\
    \end{subfigure} 
    \begin{subfigure}[t]{0.135\textwidth}
        \centering
        \includegraphics[width=\textwidth, keepaspectratio]{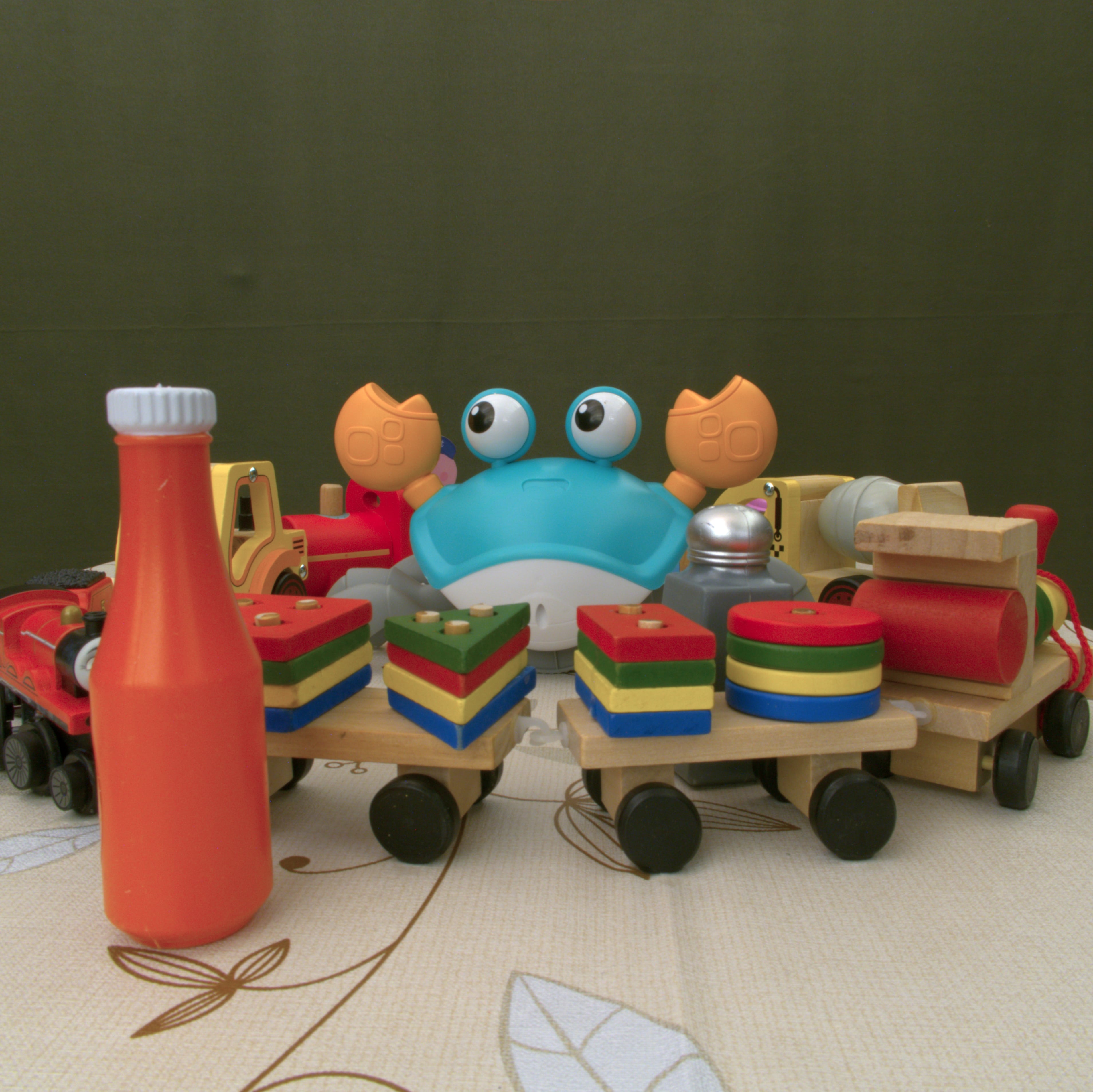}          
    \end{subfigure}    
    \hfill    
    \begin{subfigure}[t]{0.135\textwidth}
        \centering
        \includegraphics[width=\textwidth, keepaspectratio]{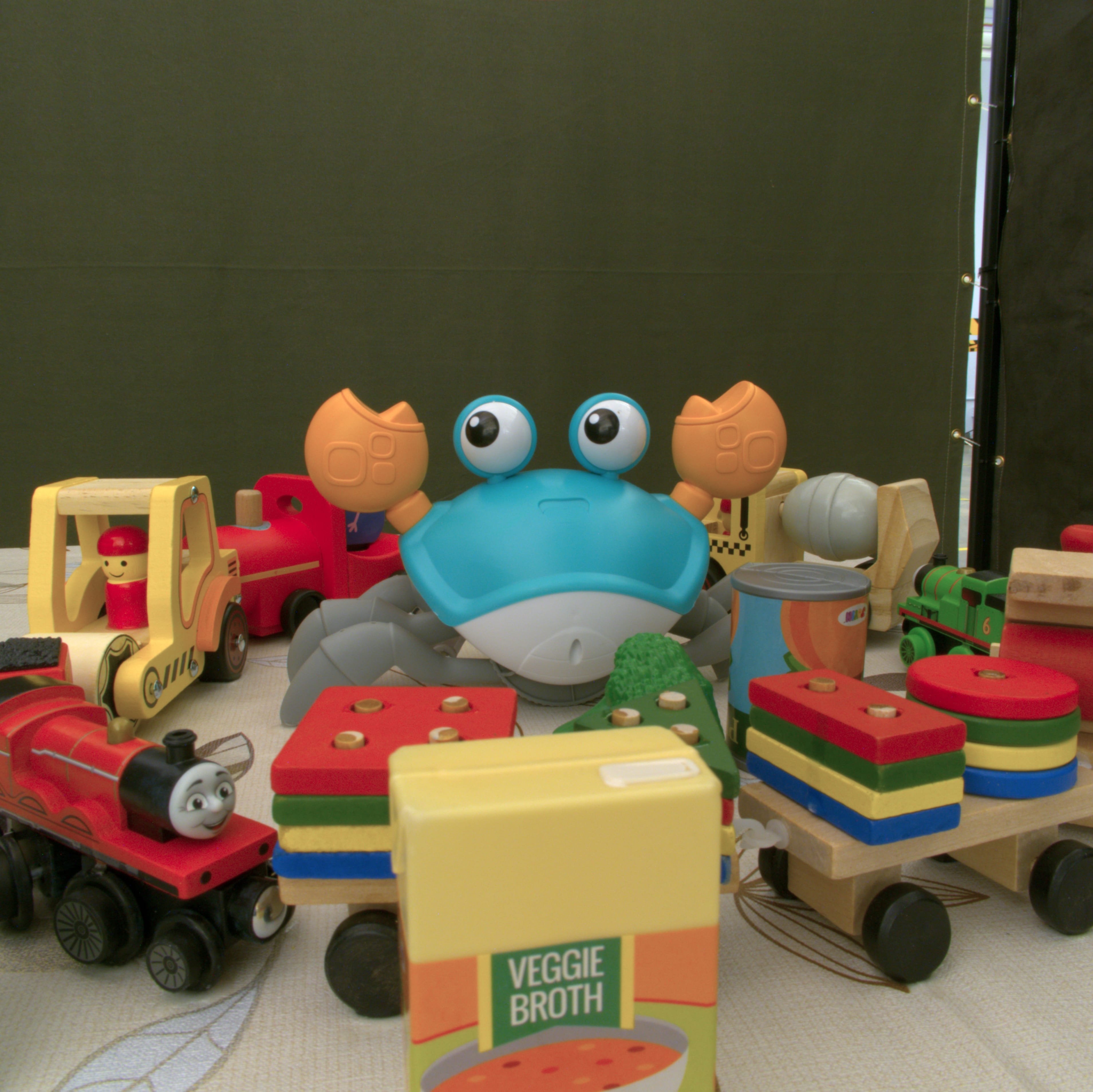}    
    \end{subfigure}    
    \hfill 
    \begin{subfigure}[t]{0.135\textwidth}
        \centering
        \includegraphics[width=\textwidth, keepaspectratio]{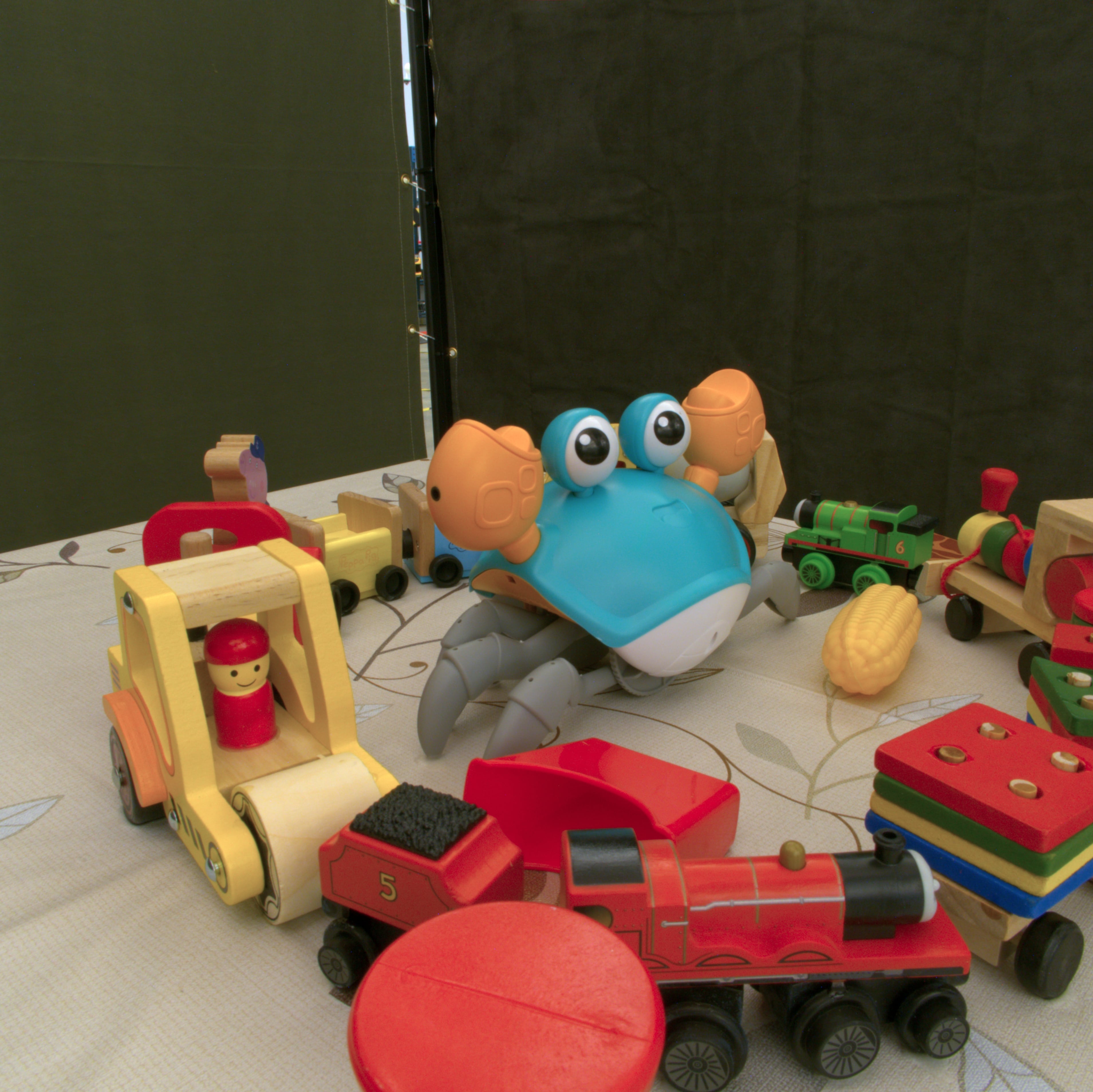}      
    \end{subfigure}   
    \hfill 
    \begin{subfigure}[t]{0.135\textwidth}
        \centering
        \includegraphics[width=\textwidth, keepaspectratio]{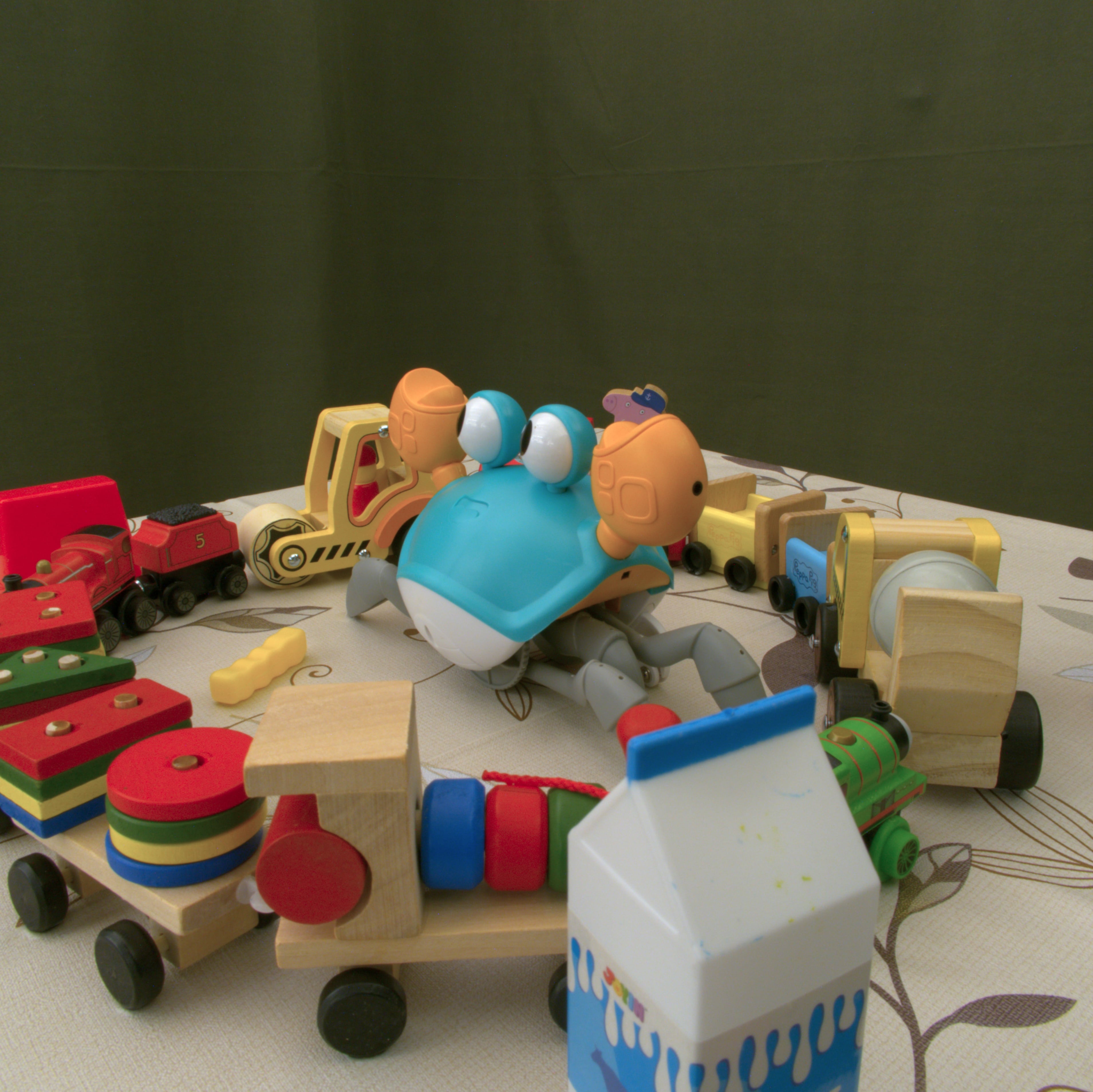}      
    \end{subfigure} 
    \hfill 
    \begin{subfigure}[t]{0.135\textwidth}
        \centering
        \includegraphics[width=\textwidth, keepaspectratio]{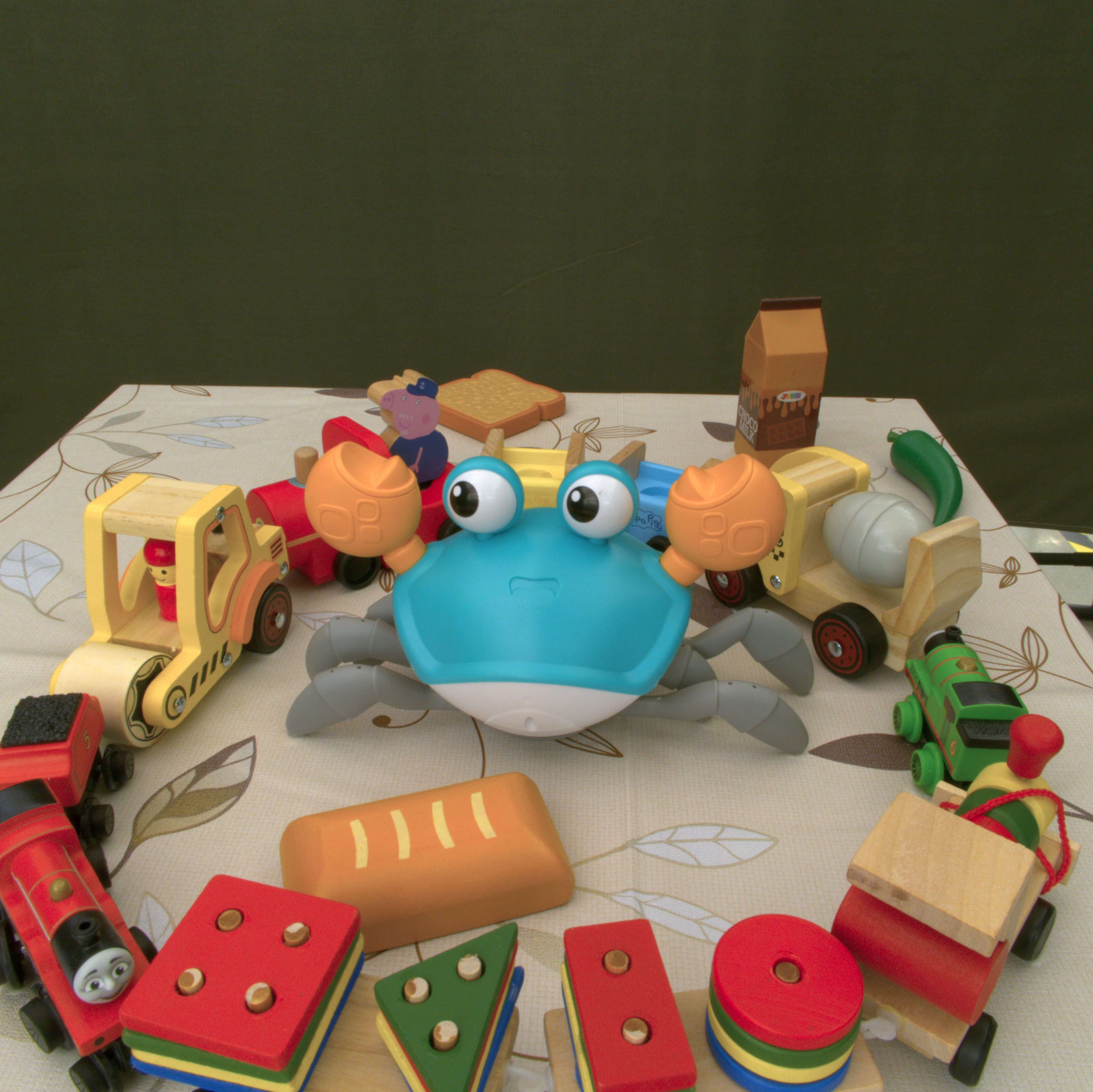}      
    \end{subfigure} 
    \hfill 
    \begin{subfigure}[t]{0.05\textwidth}
        \centering
        \vspace{-0.8cm}
        \dots  
    \end{subfigure} 
    \begin{subfigure}[t]{0.135\textwidth}
        \centering
        \includegraphics[width=\textwidth, keepaspectratio]{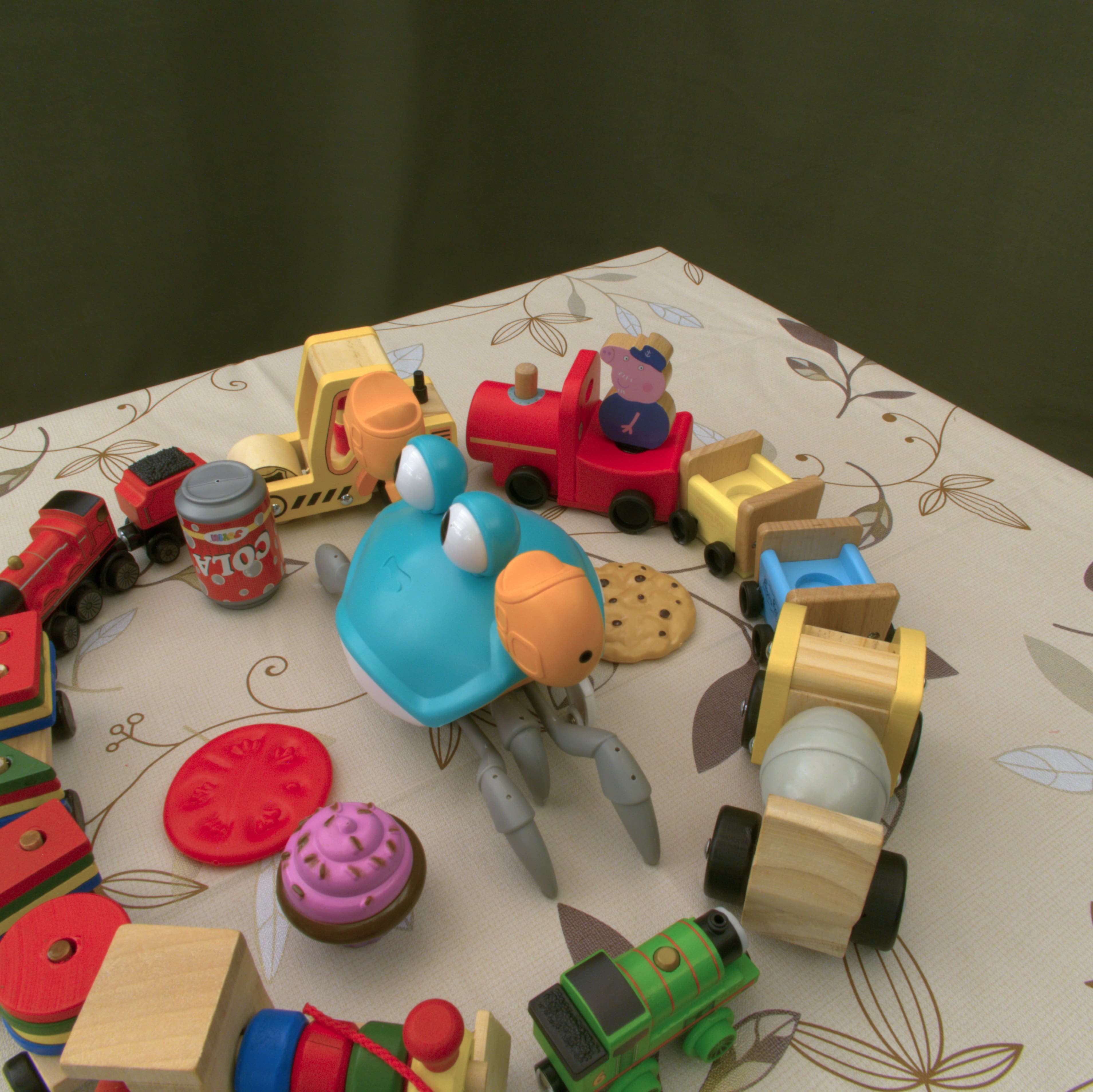}      
    \end{subfigure} 

    %\newline
    \vspace{0.04cm}
    
    \begin{subfigure}[t]{0.32\textwidth}
        \centering
        \includegraphics[width=\textwidth, keepaspectratio]{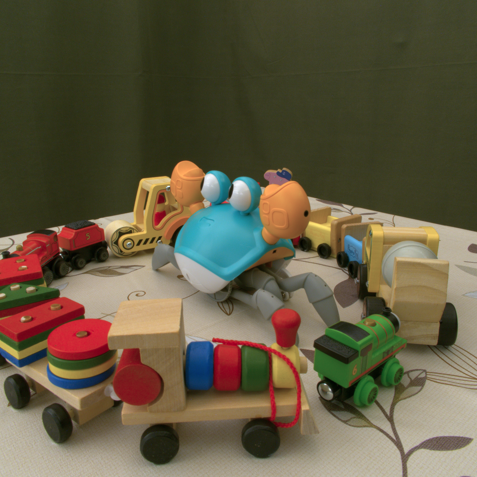}  
        \caption[]{Ground truth} 
    \end{subfigure}    
    \hfill    
    \begin{subfigure}[t]{0.32\textwidth}
        \centering
        \includegraphics[width=\textwidth, keepaspectratio]{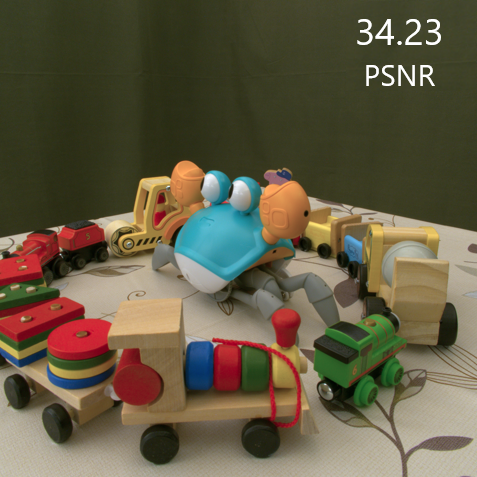}   
        \caption[]{Our method} 
    \end{subfigure}    
    \hfill 
    \begin{subfigure}[t]{0.32\textwidth}
        \centering
        \includegraphics[width=\textwidth, keepaspectratio]{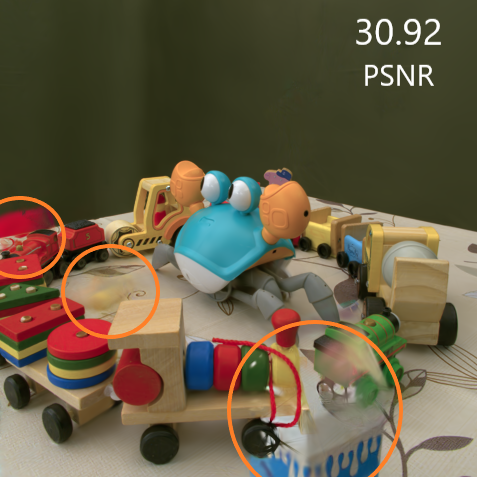}      
        \caption[]{3D Gaussian Splatting} 
    \end{subfigure}    
    \vspace{-0.1cm}
    \caption{Due to distractors in the scene 3D Gaussian Splatting creates floating artifacts in the image (highlighted with circles). Our method mitigates artifacts due to violations of the static scene assumption for Gaussian Splatting. As a key element to our approach, we optimize for semantic distractor masks simultaneous to the scene optimization, which allow us to effectively ignore distractors.}
\end{figure*}

\vspace{-0.9cm}

\begin{abstract}
3D Gaussian Splatting has shown impressive novel view synthesis results; nonetheless, it is vulnerable to dynamic objects polluting the input data of an otherwise static scene, so called distractors. Distractors have severe impact on the rendering quality as they get represented as view-dependent effects or result in floating artifacts. Our goal is to identify and ignore such distractors during the 3D Gaussian optimization to obtain a clean reconstruction. To this end, we take a self-supervised approach that looks at the image residuals during the optimization to determine areas that have likely been falsified by a distractor. In addition, we leverage a pretrained segmentation network to provide object awareness, enabling more accurate exclusion of distractors. This way, we obtain segmentation masks of distractors to effectively ignore them in the loss formulation. We demonstrate that our approach is robust to various distractors and strongly 
improves rendering quality on distractor-polluted scenes, improving PSNR by 1.86dB compared to 3D Gaussian Splatting. 

\end{abstract}

\section{Introduction}
\label{sec:intro}
Neural Radiance Fields (NeRFs)~\cite{mildenhall2021nerf} and 3D Gaussian Splatting~\cite{kerbl20233d} have shown remarkable improvements in novel view synthesis on complex scenes, enabling various tasks in the fields of virtual reality, autonomous systems, gaming or others. Given a set of input images along with camera poses, a 3D scene representation is optimized from which photo-realistic novel views can be rendered. NeRF represents the radiance and density distribution of a scene with a multi-layer perceptron (MLP) and employs differentiable volume rendering to synthesize novel views. In contrast, Gaussian Splatting represents the scene in an explicit manner, as a set of 3D Gaussians, defined by position, covariance, opacity, and spherical harmonic coefficients, which are efficiently rendered with a differentiable rasterizer, thereby enabling high-quality novel view synthesis in real-time. 

3D Gaussian Splatting and NeRF are both optimized to minimize a re-rendering loss in RGB space. This procedure relies on a static scene assumption, i.e., the images must be photometrically consistent. In real-world scenarios, however, this assumption of a perfectly static scene is hardly ever fulfilled. Even with careful scene capture, the recordings often contain dynamics, such as lighting changes, moving shadows, or any unforeseeable moving objects or persons, e.g., tourists near a captured landmark. We refer to such undesired dynamic observations as distractors. 
Ignoring this problem severely degrades the optimized scene representation and the rendering quality, resulting in floating artifacts and blurriness (\cref{fig:c_segmentation}). 
At the same time, the removal of distractors from a dataset as a post-processing step is non-trivial due to the variety of potential distractors. Clearly, manual pixel-wise annotation of distractors is impractical. Finally, for making Gaussian Splatting widely adopted and applicable to in-the-wild settings, the data capture has to remain simple with little effort. 
Therefore, it is highly desirable to increase the robustness of 3D Gaussian Splatting to distractors to be able to obtain clean reconstructions even from imperfect data. 
RobustNeRF~\cite{sabour2023robustnerf} provides a solution for handling distractors in NeRF using a robust loss that computes distractor masks through iteratively reweighted least squares. Applying this approach to Gaussian Splatting, however, causes too aggressive masking and reduced performance (\cref{sec:experiments}). 
Hence, we take a different approach and learn flexible neural decision boundaries to distinguish between distractors and static scene content. 

Our work simultaneously optimizes for distractor masks to support the static scene reconstruction. To this end, we leverage image residuals in the training process and apply different transformations to obtain local smoothness and distractor contiguity. Using a neural classifier to classify distractor pixels, we compute a first semantic distractor mask. We refine these segmentation masks and establish object awareness using the object segmentation mask from SegmentAnything~\cite{kirillov2023segment}. At the same time, our method remains independent from the type of distractors and can handle arbitrary distractors. We evaluate our approach on challenging distractor-polluted scenes~\cite{sabour2023robustnerf} and obtain remarkable improvements over 3D Gaussian Splatting and RobustNeRF by 1.9dB and 4.3dB in PSNR, respectively. 

In summary, we provide the following contributions:
 \begin{itemize}
    \item We introduce distractor masks by optimizing a neural decision boundary based on image residuals to effectively track and exclude distractors during 3D Gaussians optimization. 
    \item We propose to leverage a pretrained segmentation network to enhance the distractor masks, making them object aware for more accurate exclusion of distractors. 
\end{itemize}

\section{Related Work}

\subsubsection{Neural Radiance Fields and 3D Gaussian Splatting} Neural Radiance Fields (NeRF)~\cite{mildenhall2021nerf} achieve outstanding novel view synthesis performance. The NeRF scene representation is realized as an implicit function, where a MLP maps a 3D position and viewing direction to radiance and density. Differentiable volume rendering combines radiance and density along target camera rays to produce pixel colors in the output view. 
The scene representation is optimized on a set of posed input images by minimizing a photometric loss. At inference time, novel views can be rendered from arbitrary view points. 
Follow-up works have extended NeRF in many directions, for instance towards alternative scene representations, e.g., voxel grids~\cite{fridovich2022plenoxels,liu2020neural,sun2022direct}, decomposed tensors~\cite{chan2022efficient,kplanes_2023,tensorf}, or hash maps~\cite{mueller2022instant} to increase optimization and rendering speed. Various extensions also exist towards novel view synthesis on dynamic scenes~\cite{pumarola2021d,tretschk2021non,park2021nerfies,gafni2021dynamic}
, where typically a deformation field is optimized in addition to a canonical NeRF. A bit less explored, but still highly relevant is the application of NeRF to distractor-polluted scenes to which we dedicate an individual paragraph below.  

3D Gaussian Splatting \cite{kerbl20233d} is a recent method for novel view synthesis. In contrast to NeRF, a set of multivariate Gaussians with parameters such as position, covariance, opacity and appearance, is optimized as a scene representation. 3D Gaussian Splatting leverages a differentiable rasterizer that efficiently renders the Gaussians which allows real-time rendering, while at the same time outperforming NeRF in image quality metrics. Recent advances in Gaussian Splatting mainly focus on improving image quality in different distractor-free settings, namely dynamic scenes \cite{wu20234d,yu2023cogs,yang2023real,liang2023gaufre} and static scenes \cite{yugay2023gaussian,matsuki2023gaussian,yan2023gs,keetha2023splatam}. Other research directions tackle data efficiency using depth information \cite{xiong2023sparsegs,zhu2023fsgs,chung2023depth}.
Furthermore, Gaussian Splatting is also used to improve generative models like \cite{tang2023dreamgaussian,chen2023text}.

\subsubsection{Handling Distractors.} There are several approaches to handle distractors in general. In settings where distractors are known to be in specific classes we can employ a pretrained semantic segmentation model to remove distractors \cite{rematas2022urban,tancik2022block}. The main problem with this approach are distractors that do not belong to any known class such as shadows. Another approach is to exploit the time dependency of the images to classify static and dynamic (i.e., distractor) objects in the scene \cite{wu2022d}. The problem with this method is that it requires time dependency, which typically is not available in the multi-view reconstruction setting. Nonetheless, we take inspiration from the usage of pretrained semantic segmentation networks and leverage SegmentAnything~\cite{kirillov2023segment} to provide object awareness in our distractor mask optimization, while at the same time maintaining flexibility to arbitrary distractor categories. 

\subsubsection{Robust Methods for NeRF.} For traditional NeRFs, a promising method to mitigate problems with distractors, is to use a robust loss function \cite{sabour2023robustnerf}. RobustNeRF \cite{sabour2023robustnerf} is a robust technique for dealing with distractors. It computes a segmentation mask to ignore distractors in the loss during training. The masks are calculated in each iteration for each image through iteratively reweighted least squares. The idea is that the masks converge over time to the true distractor segmentation.
Another approach is to use data-driven priors to remove artifacts from the image \cite{warburg2023nerfbusters}.

Up to now, there has been no work on distractor handling for 3D Gaussian Splatting.

\section{Method}
\label{sec:methods}
Given a set of input images from a scene with the corresponding camera poses, Robust 3D Gaussian Splatting optimizes for a clean 3D scene representation which ignores any distractors that may be present in the input. To mitigate the problems caused by distractors, we present an approach to identify and track distractors simultaneous to the scene optimization. 
First, we compute raw masks from image residuals and process them for better spatial smothness and contiguous local support (\cref{ssec:raw_masks}). Then, we apply a logistic regression learning to distinguish distractor from non-distractor pixels (\cref{ssec:neural_decision_boundary}). Ultimately, we intersect the resulting mask with object segmentation masks from a pretrained network to enable object awareness (\cref{ssec:object_awareness}). The resulting distractor mask is used in the loss formulation, such that image parts containing distractors are effectively ignored in the optimization (\cref{fig:training_cycle}). 

\subsection{Raw Mask Generation}
\label{ssec:raw_masks}
Similar to \cite{sabour2023robustnerf}, we build distractor masks during the optimization. However, instead of hard thresholds, we use a logistic regression to flexibly learn thresholds. Furthermore, we also compute a mask for each image channel, which increases performance. 

We first center the residuals from the last iteration $\epsilon(\textbf{R}):=|\textbf{R}^{(i)}_\text{GT} - \textbf{R}^{(i)}_\text{render}|$ for all pixels in the image $\textbf{R}$ of each channel $i \in C$ using the median:
\begin{equation}
    \hat{\epsilon}(\textbf{R})^{(i)} = \epsilon(\textbf{R})^{(i)} - \text{median}\{\epsilon(\textbf{R})^{(i)}\}.
\end{equation}
For robustness, we use the median for centering since the residuals have high variance. Then, we apply a $3\times3$ box kernel to capture local smoothness and obtain a better local continuity: 
\begin{equation}
    \label{eq:box_kernel}
    \mathcal{\omega}_1(\textbf{R})^{(i)} = \hat{\epsilon}(\textbf{R})^{(i)} \circledast \mathcal{B}_{3 \times 3}.
\end{equation}
Next, we compute the value of whole $8 \times 8$ patches in a $16 \times 16$ neighborhood. This allows us to capture a more contiguous behavior of distractors. We define
\begin{equation}
    \omega_2\left(\mathcal{R}_8(\textbf{R})\right)^{(i)} = \frac{1}{16^2} \sum_{\textbf{s} \in \mathcal{R}_{16}(\textbf{R})} \omega_1(\textbf{s})^{(i)},
\end{equation}
where $\mathcal{R}_N(\textbf{R})$ describes the $N \times N$ neighbourhood around \textbf{R}. 
In the next step, we aggregate all information to obtain a better approximation
\begin{equation}
    \omega_3(\textbf{R})^{(i)} = \hat{\epsilon}(\textbf{R})^{(i)} + \omega_1(\textbf{R})^{(i)} + \omega_2(\textbf{R})^{(i)}.
    \label{eq:raw_mask}
\end{equation}
\subsection{Neural Decision Boundary}
\label{ssec:neural_decision_boundary}
Now, we learn the decision boundary using a logistic regression 
\begin{equation}
    \label{eq:log_reg}
    \hat{\mathcal{W}}(\textbf{R}) = \sigma\left(\textbf{W} \omega_3(\textbf{R}) + b \right),
\end{equation}
where $\textbf{W}$ and $b$ are learned parameters and $\sigma(\cdot)$ is the sigmoid function. Note that we apply the logistic regression pixel-wise and aggregate the channel after using the median.
\begin{figure*}[!t]
    \centering
    \label{fig:mask_comparison}
    \centering
    \begin{subfigure}[t]{0.23\textwidth}
        \centering
        \includegraphics[width=\textwidth, keepaspectratio, frame]{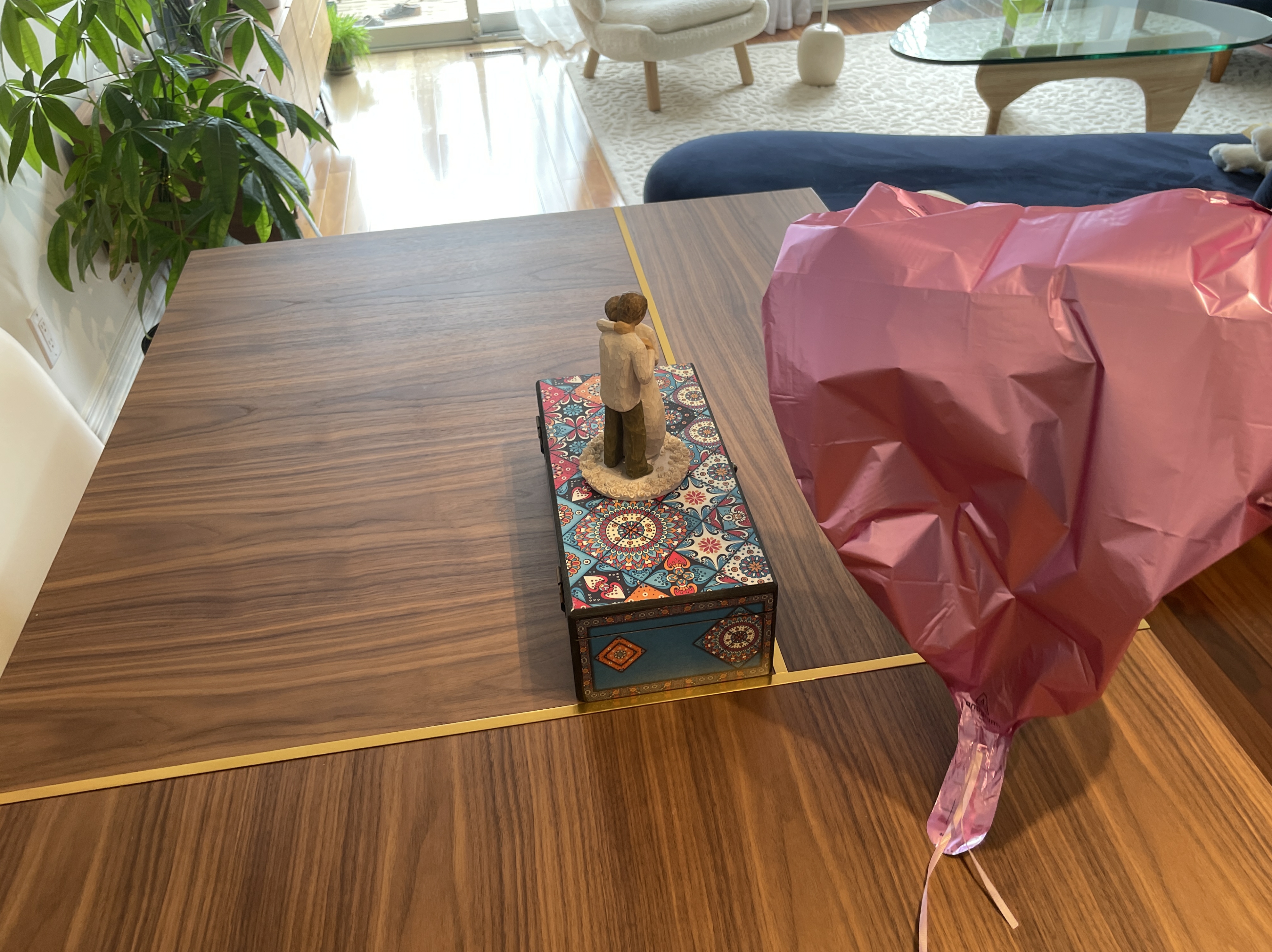}  
        \caption[]{Ground truth} 
        \label{fig:c_gound_truth}
    \end{subfigure}    
    \hfill    
    \begin{subfigure}[t]{0.23\textwidth}
        \centering
        \includegraphics[width=\textwidth, keepaspectratio, frame]{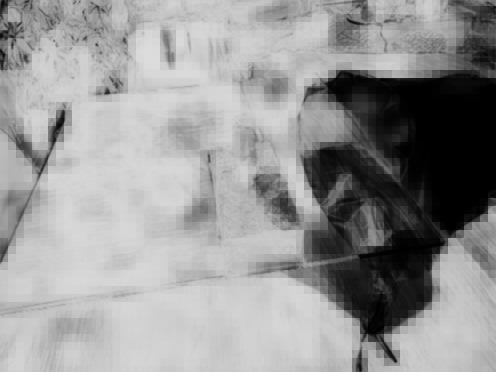}      
        \caption[]{Raw distractor mask} 
        \label{fig:c_raw_mask}
    \end{subfigure}    
    \hfill
    \begin{subfigure}[t]{0.23\textwidth}
        \centering
        \includegraphics[width=\textwidth, keepaspectratio, frame]{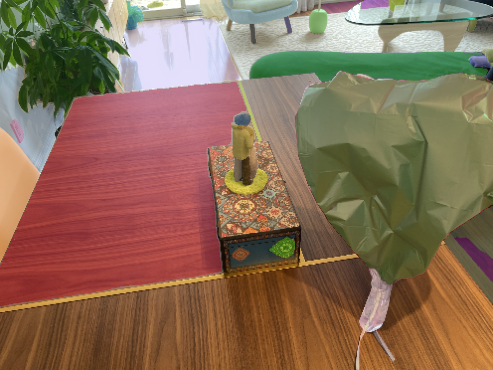}      
        \caption[]{Segmentation from SAM \cite{kirillov2023segment}} 
        \label{fig:c_segmentation}
    \end{subfigure}    
    \hfill    
    \begin{subfigure}[t]{0.23\textwidth}
        \centering
        \includegraphics[width=\textwidth, keepaspectratio, frame]{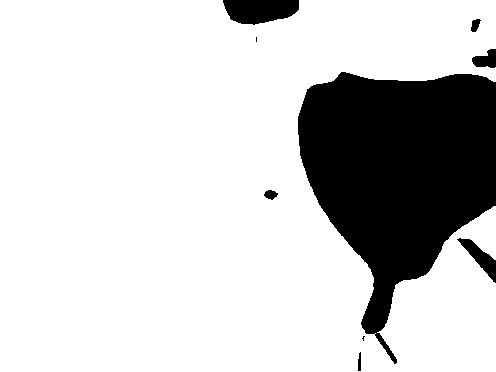}     
        \caption[]{Intersected mask} 
        \label{fig:c_intersection}
    \end{subfigure}
    \caption{Given the ground truth (\cref{fig:c_gound_truth}) and the rendered image we can calculate the raw distractor mask (see \cref{eq:log_reg}). Next, we intersect the raw distractor mask with the object masks from SegmentAnything (\cref{fig:c_segmentation}). The intersected mask (\cref{fig:c_intersection}) is then used in the loss.}
\end{figure*}

The idea is that the higher the mask value of a pixel, the more likely it is to be a distractor. This is the case, because we calculate the masks using the residuals. Distractors cause artifacts in the renderings and pixels with artifacts have high residuals, because the artifacts do not match the target image. Another cause of high residuals is that the model has not been trained enough. However, since we are using a dynamic decision classifier that can change over time, we can mitigate this problem. 
For simplicity reasons, we define 
\begin{equation}
    \hat{\mathcal{W}}^c := 1 - \hat{\mathcal{W}}.
\end{equation}
Note that in $ \hat{\mathcal{W}}$, distractors are labeled as 0 so they can be directly ignored in the loss.

We train the logistic regression by calculating a custom mask loss using the Gaussian Splatting loss of a rendered image $\mathcal{L}_{\text{GS}}(\textbf{R}, \textbf{G})$, where $\textbf{G}$ is the ground truth image. We define the mask loss using the corresponding mask $\hat{\mathcal{W}}$ for the image \textbf{R} as
\begin{equation}
    \label{eq:mask_loss}
    \mathcal{L}_{\text{mask}}(\textbf{R}, \textbf{G}, \hat{\mathcal{W}}) = \mathcal{L}_{\text{GS}}(\textbf{R} \circ \hat{\mathcal{W}}, \textbf{G} \circ \hat{\mathcal{W}}) + \frac{\lambda}{mn} \sum_i^m \sum_j^n \hat{\mathcal{W}}_{i,j}^c
\end{equation}
where $\lambda$ is the regularization strength, $m, n$ is the image height and width and $\circ$ is the Hadamard product. The main challenge of this approach is that we do not have a ground truth mask. Because of that, we compute the mask's impact on the Gaussian Splatting loss. The trivial solution for the logistic regression is to classify every pixel as a distractor since we ignore the distractor pixel in the loss and thus obtain a loss of $0$. Therefore, we regularize the loss by the proportion of distractor pixels in the mask. After the mask loss calculation, we round our mask
\begin{equation}
    \bar{\mathcal{W}} := \text{round} ( \hat{\mathcal{W}}).
\end{equation}
The idea of the neural decision boundary is to dynamically find a threshold value to classify pixels as distractors. Instead of a fixed threshold we used a logistic regression because the logistic regression adapts itself during training and is able to shift the threshold value throughout the training process.
In the following, we refer to the non-binary masks as raw masks.

\subsection{Establishing Object Awareness}
\label{ssec:object_awareness}
We can see in \cref{fig:c_raw_mask} that the raw masks capture the coarse structure of the distractor. However, we can also see that the distractor as a whole is not completely correctly classified, and some non-distractor parts also have high mask values. We propose to use only whole objects with an intersection ratio of more than 40\% between the object and the mask. We define the object mask set for the image \textbf{R} as $\text{M}(\textbf{R})$. This set contains a segment mask for every object in the image, i.e., a matrix where every pixel is $1$ if it belongs to the object. That means $|M(\textbf{R})| = \#\text{objects in the image}$. We obtain the object segmentations from SegmentAnything \cite{kirillov2023segment}. An example of the combined segments can be found in \cref{fig:c_segmentation}. Now, we can define the intersection of each object with the mask as
\begin{equation}
    \label{eq:intersections}
    \mathcal{I}(M(\textbf{R})) = \left\{m \in M(\textbf{R}) | J\left(m \circ \bar{\mathcal{W}}^c(\textbf{R}), m\right) > 0.4\right\}.
\end{equation}
where $J(\cdot, \cdot)$ is the Jaccard index and $\bar{\mathcal{W}}$ the rounded mask for image $\textbf{R}$. Intuitively, we label a whole object as a distractor if enough pixels in this object are classified as a distractor in the mask computed from the residuals. The intersected mask as in \cref{fig:c_intersection} is defined as follows:
\begin{equation}
    \label{eq:sum_mask}
    \mathcal{W}(\textbf{R}) = \sum_{m \in \mathcal{I}(M(\textbf{R}))} (1-m).
\end{equation}
Note that we switch between $\bar{\mathcal{W}}^c$ and $\bar{\mathcal{W}}$ to maintain the convention that distractor pixels are labeled as 0, 
hence the subtraction in \cref{eq:sum_mask}.

\begin{figure*}[!hb]
    \centering
    \includegraphics[scale=1]{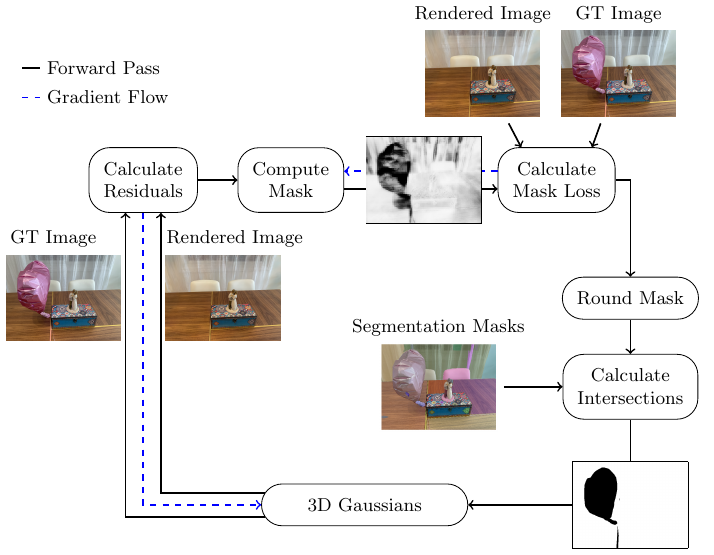}
    \caption{The first step is to calculate the residuals using the ground truth image and the rendered image. Next, we compute the mask as described in \cref{eq:log_reg} using the neural decision boundary. Then, we calculate the mask loss from \cref{eq:mask_loss} using the computed mask and propagate back to learn the logistic regression. After that, we intersect the mask with the segmentation masks from Segment Anything and round the masks. The ground truth and the rendered image are multiplied element-wise with the distractor mask and used in the Gaussian Splatting training.}
    \label{fig:training_cycle}
\end{figure*}

Using the intersected masks, we can now ignore the distractors in the loss of the Gaussian Splatting optimization. 
The complete mask generation process is summarized in \cref{fig:training_cycle}.

The computation of SAM during training does not significantly influence the  runtime. Averaged over different scenes, we observe a runtime increase of less than 1\%.

\section{Experiments}
\label{sec:experiments}
\subsection{Dataset}
Following RobustNeRF~\cite{sabour2023robustnerf}, we evaluate our method on the same four scenes of their released dataset. This dataset was captured to facilitate novel view synthesis on challenging scenes which contain various distractors. 
Each scene contains training images with distractors and test images that show the scene without any distractors from different camera poses. There is no temporal correlation between images in the dataset. As in the RobustNeRF paper, we also downscaled the images by a factor of 8 to ensure consistent comparisons.
The scenes can be described as follows:
\begin{itemize}
    \item Statue (225 train/19 test): A wooden statue on a painted box. Distractor: a red balloon.
    \item Yoda (63/202): A stuffed Baby Yoda and toy animals. Distractors: different household items.
    \item And-bot (122/19): Two Android figurines on a board game box. Distractors: wooden figures.
    \item Crab (72/72): A crab figure surrounded by a toy train. Distractors: different household items.
\end{itemize}
Furthermore, we compare our method on a real life scene, see supplementary material.

\subsection{Baseline Methods and Ablations}
We compare against 3D Gaussian Splatting~\cite{kerbl20233d}, which our method builds upon, as well as to RobustNeRF~\cite{sabour2023robustnerf}, the most recent and strongest baseline for novel view synthesis in presence of distractors. We further provide results on the straight-forward combination of 3D Gaussian Splatting with the robust loss from RobustNeRF. . 
To analyze the effectiveness of our added components, we conduct ablation experiments for the neural decision boundary optimization, as well as the object awareness using a pretrained segmentation network. For all methods we use the same training configuration as described in the supplementary material.
All in all, we conduct comparisons on the following six approaches. 
\begin{itemize}
    \item \textbf{Gaussian Splatting}: The unmodified implementation of 3D Gaussian Splatting as introduced by \cite{kerbl20233d}.
    \item \textbf{RobustNeRF (GS)}: A version of RobustNeRF implemented for 3D Gaussian Splatting.
    \item \textbf{RobustNeRF (NeRF)}: The original version of RobustNeRF \cite{sabour2023robustnerf} using mip-NeRF 360 \cite{barron2022mip}.
    \item \textbf{Raw Masks + Segmentation (w/o Neural)}: Sequentially adds the segmentation overlap improvements after calculating the raw masks but does not use a logistic regression.
    \item \textbf{Raw Masks + Neural Decision Boundary (w/o Segmentation)}: Sequentially adds a trainable neural decision boundary (logistic regression) after calculating the raw mask but does not use segmentation.
    \item \textbf{Raw Masks + Neural Decision Boundary + Segmentation (Robust Gaussian Splatting)}: Our complete method as described in \cref{sec:methods} and shown in \cref{fig:training_cycle}. 
\end{itemize}
We evaluate in terms of three visual quality metrics: peak signal-to-noise ratio (PSNR), SSIM~\cite{wang2004image}, and LPIPS~\cite{zhang2018unreasonable}.
 
\begin{table*}[h]
\vspace{-0.7cm}
    \centering
        \caption{Quantitative comparison to baselines and ablated versions averaged over multiple scenes. The individual metrics are provided in the supplementary material. Values for the RobustNeRF (NeRF) implementation are taken from \cite{sabour2023robustnerf}. }
    \begin{tabularx}{\textwidth}{|>{\centering\arraybackslash}X||>{\centering\arraybackslash}X|>{\centering\arraybackslash}X|>{\centering\arraybackslash}X|>{\centering\arraybackslash}X|>{\centering\arraybackslash}X|>{\centering\arraybackslash}X|}
        \hline
        Metric&Gaussian Splatting&Robust NeRF (GS)&Robust NeRF (NeRF)&w/o Segmentation&w/o Neural&Robust Gaussian Splatting (\textbf{Ours})\\
        \hhline{|=|=|=|=|=|=|=|}
        \hline
        SSIM \(\uparrow\)&0.87253&0.8579&0.76&0.8842&0.8273&\textbf{0.8875}\\
        PSNR \(\uparrow\)&26.53&24.09&26.06&27.95&23.28&\textbf{28.39}\\
        LPIPS \(\downarrow\)&0.1568&0.1773&0.25&0.1363&0.1698&\textbf{0.1321}\\
        \hline

    \end{tabularx}
\vspace{-0.7cm}
    \label{tab:quant_results}
\end{table*}

\subsection{Quantitative Comparisons}
The quantitative result of the evaluation can be seen in~\cref{tab:quant_results}. As the table shows, our method outperforms all other models. We averaged the performance of the models over all scenes. The detailed per scene performance is provided in the supplementary material.

Additionally, we performed an ablation test to investigate if using three color channels as input to the mask calculation improves results over using their norm. The test showed that they provide a small but consistent improvement.

We analyzed the behaviour of Robust Gaussian Splatting on clean data. We found that it performs fairly even to normal Gaussian Splatting on clean images. For detailed values, see supplementary material.

\subsection{Qualitative Comparisons}
\begin{figure*}[!hp]
    \centering 
    \begin{subfigure}[t]{0.24\textwidth}
        \centering
       \includegraphics[width=\textwidth,keepaspectratio]{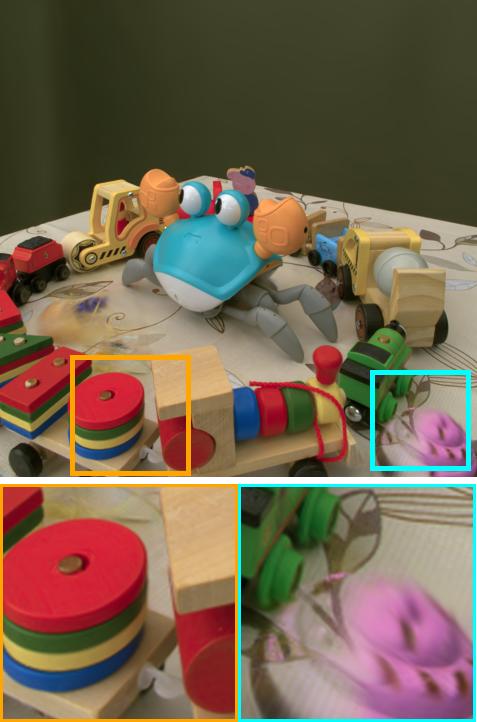}        
    \end{subfigure} 
    \hfill
    \begin{subfigure}[t]{0.24\textwidth}
        \centering
       \includegraphics[width=\textwidth,keepaspectratio]{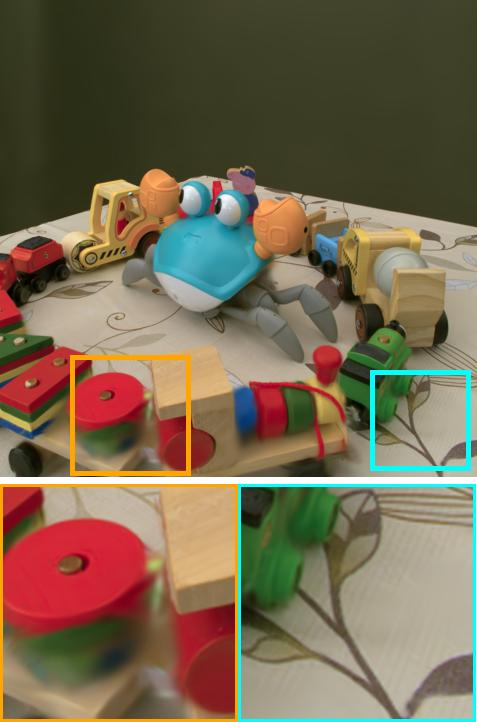}     
    \end{subfigure}  
    \hfill
    \begin{subfigure}[t]{0.24\textwidth}
        \centering
       \includegraphics[width=\textwidth,keepaspectratio]{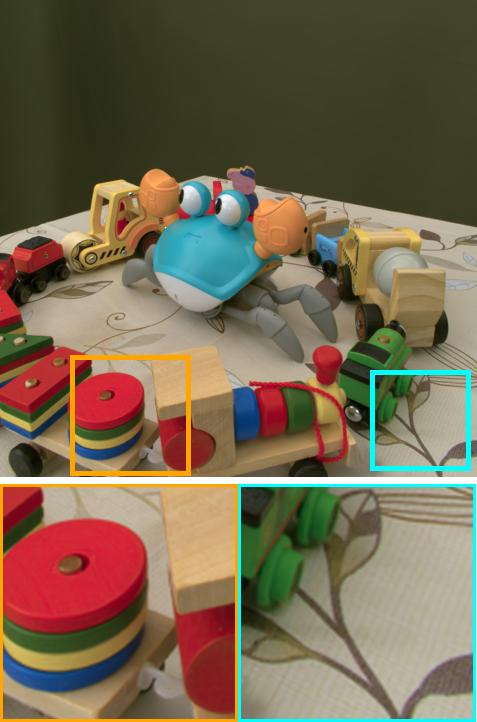}       
    \end{subfigure} 
    \hfill
    \begin{subfigure}[t]{0.24\textwidth}
        \centering
       \includegraphics[width=\textwidth,keepaspectratio]{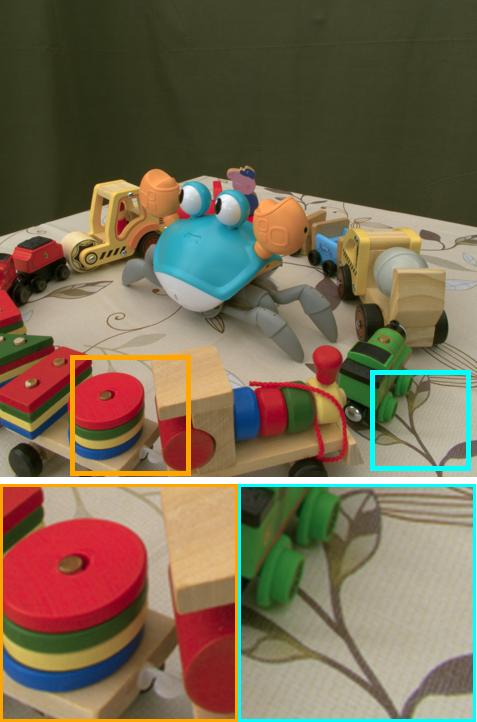}       
    \end{subfigure} 

    \vspace{0.2cm}
 
    \begin{subfigure}[t]{0.24\textwidth}
        \centering
       \includegraphics[width=\textwidth,keepaspectratio]{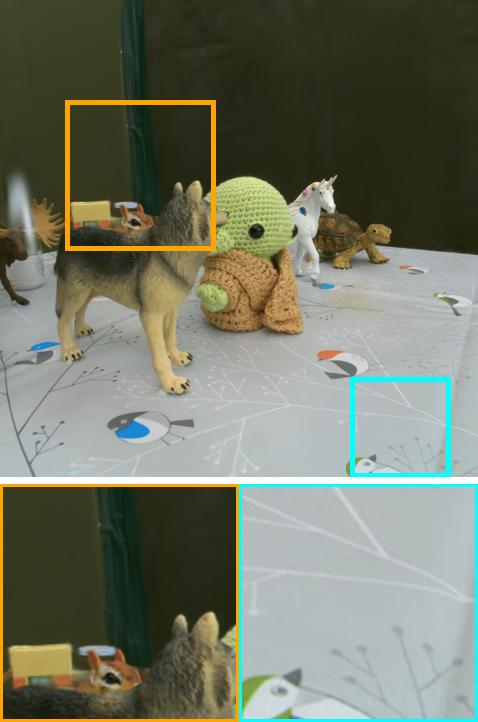}        
    \end{subfigure} 
    \hfill
    \begin{subfigure}[t]{0.24\textwidth}
        \centering
       \includegraphics[width=\textwidth,keepaspectratio]{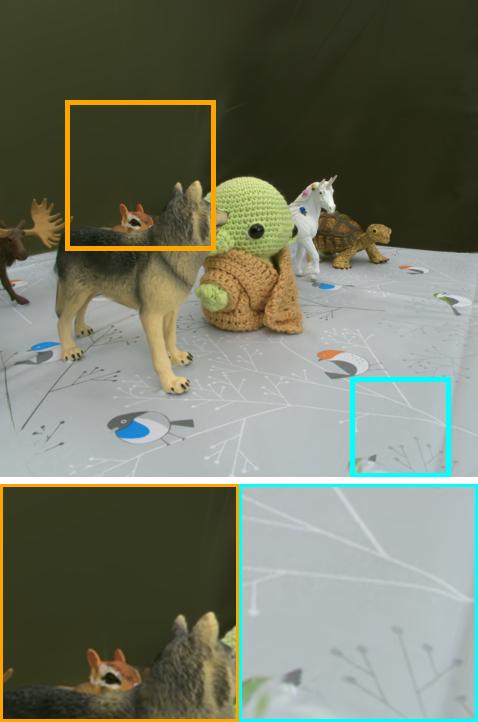}     
    \end{subfigure}  
    \hfill
    \begin{subfigure}[t]{0.24\textwidth}
        \centering
       \includegraphics[width=\textwidth,keepaspectratio]{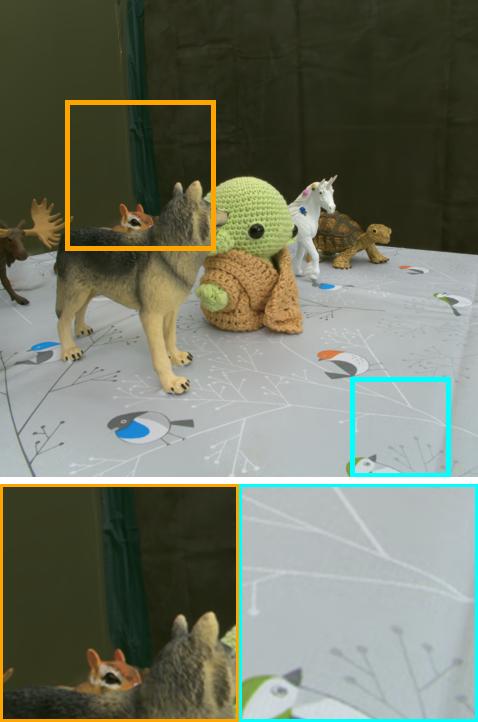}       
    \end{subfigure} 
    \hfill
    \begin{subfigure}[t]{0.24\textwidth}
        \centering
       \includegraphics[width=\textwidth,keepaspectratio]{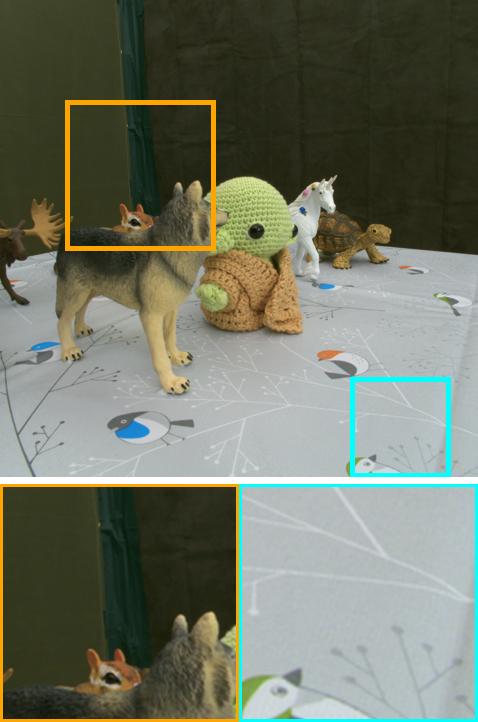}    
    \end{subfigure}  

    \vspace{0.2cm}
 
    \begin{subfigure}[t]{0.24\textwidth}
        \centering
       \includegraphics[width=\textwidth,keepaspectratio]{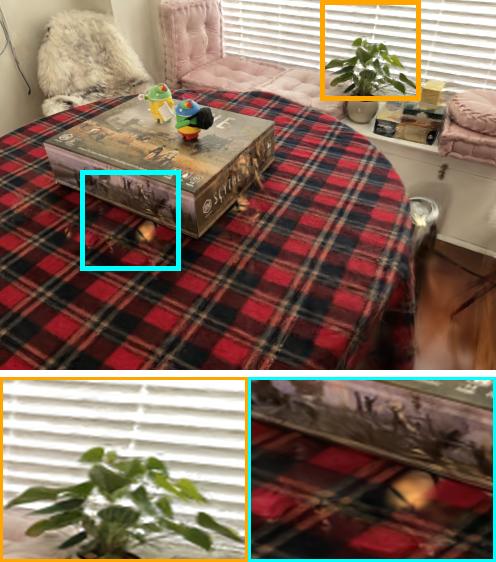}        
    \end{subfigure} 
    \hfill
    \begin{subfigure}[t]{0.24\textwidth}
        \centering
       \includegraphics[width=\textwidth,keepaspectratio]{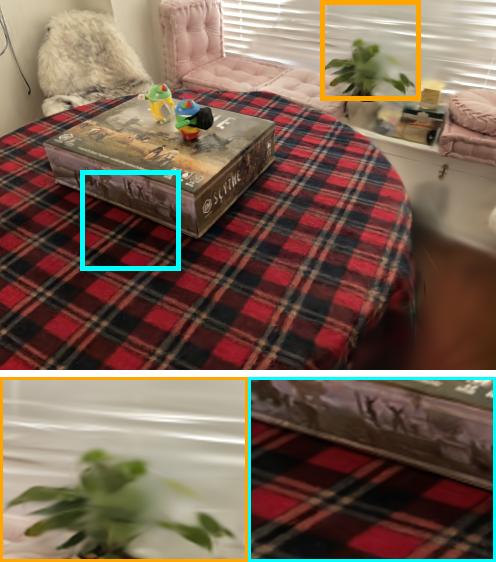}     
    \end{subfigure}  
    \hfill
    \begin{subfigure}[t]{0.24\textwidth}
        \centering
       \includegraphics[width=\textwidth,keepaspectratio]{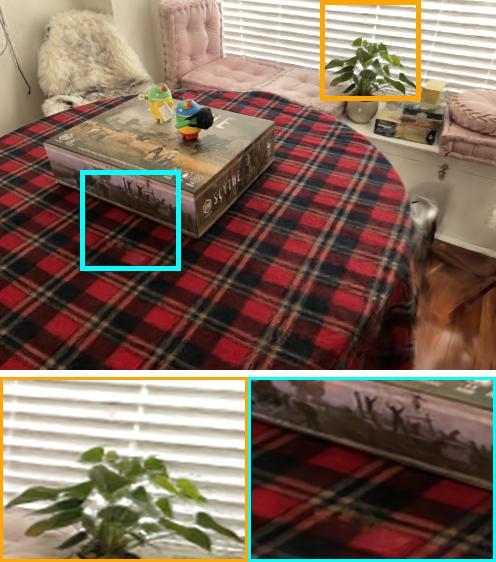}       
    \end{subfigure} 
    \hfill
    \begin{subfigure}[t]{0.24\textwidth}
        \centering
       \includegraphics[width=\textwidth,keepaspectratio]{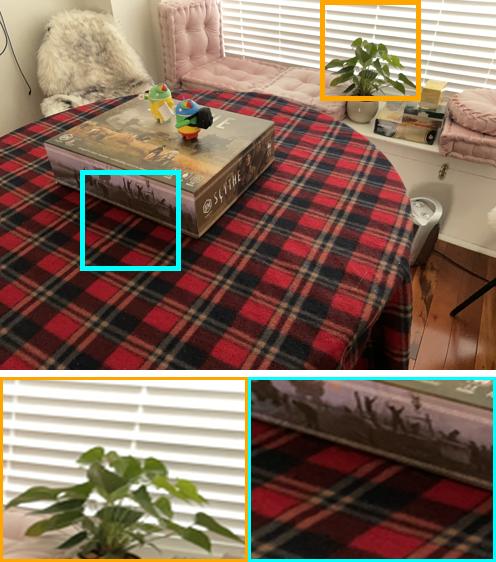}    
    \end{subfigure} 

    \vspace{0.2cm}
 
    \begin{subfigure}[t]{0.24\textwidth}
        \centering
       \includegraphics[width=\textwidth,keepaspectratio]{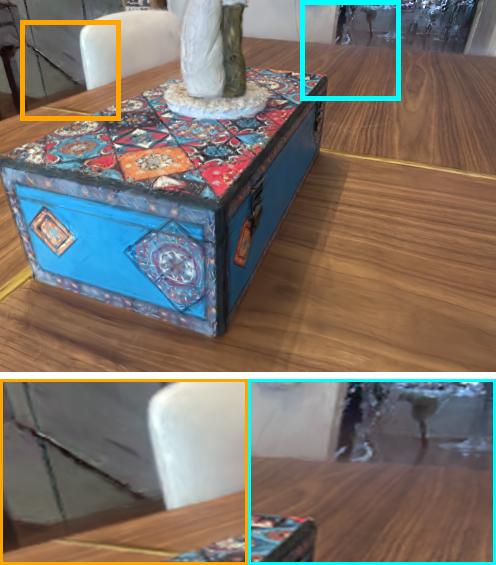}        
    \end{subfigure} 
    \hfill
    \begin{subfigure}[t]{0.24\textwidth}
        \centering
       \includegraphics[width=\textwidth,keepaspectratio]{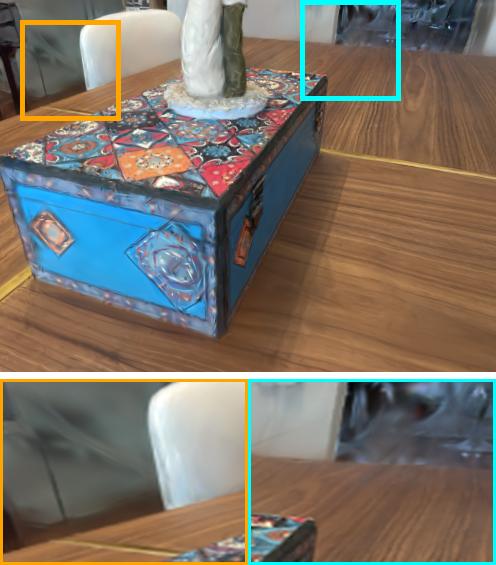}     
    \end{subfigure}  
    \hfill
    \begin{subfigure}[t]{0.24\textwidth}
        \centering
       \includegraphics[width=\textwidth,keepaspectratio]{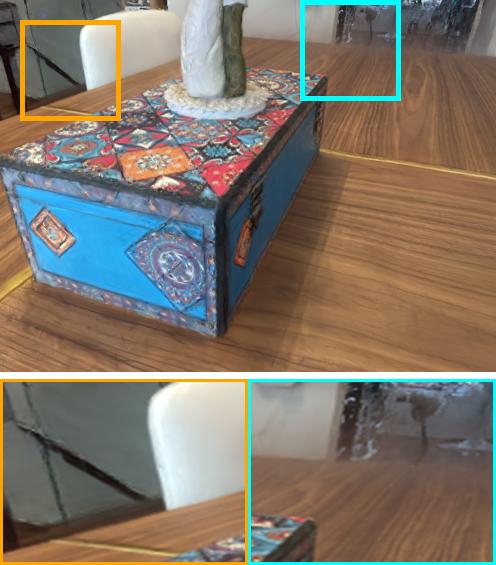}       
    \end{subfigure} 
    \hfill
    \begin{subfigure}[t]{0.24\textwidth}
        \centering
       \includegraphics[width=\textwidth,keepaspectratio]{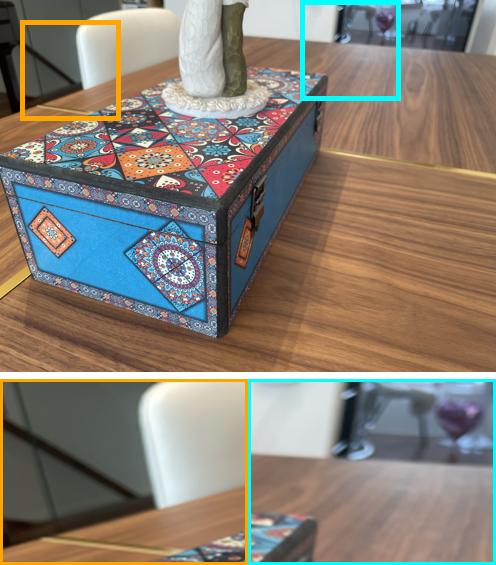}    
    \end{subfigure}  
    
    \begin{subfigure}[t]{0.24\textwidth}
        \centering
        Gaussian Splatting
    \end{subfigure}    
    \hfill  
    \begin{subfigure}[t]{0.24\textwidth}
        \centering
        RobustNeRF (GS)
    \end{subfigure}    
    \hfill   
    \begin{subfigure}[t]{0.24\textwidth}
        \centering  
        Robust Gaussian Splatting (\textbf{Ours})
    \end{subfigure}    
    \hfill
    \begin{subfigure}[t]{0.24\textwidth}
        \centering
        Ground truth 
    \end{subfigure}   
    
    \caption{Example comparison of qualitative results for all scenes from held-out test views. Robust Gaussian Splatting is most effective in ignoring distractors while maintaining a good background and general image quality. For more baseline comparisons see the supplementary material.}
    \label{fig:qual_2}
\end{figure*}

\begin{figure}[h]
    \centering  
    \begin{subfigure}[t]{0.24\textwidth}
        \centering
        \includegraphics[width=\textwidth, keepaspectratio]{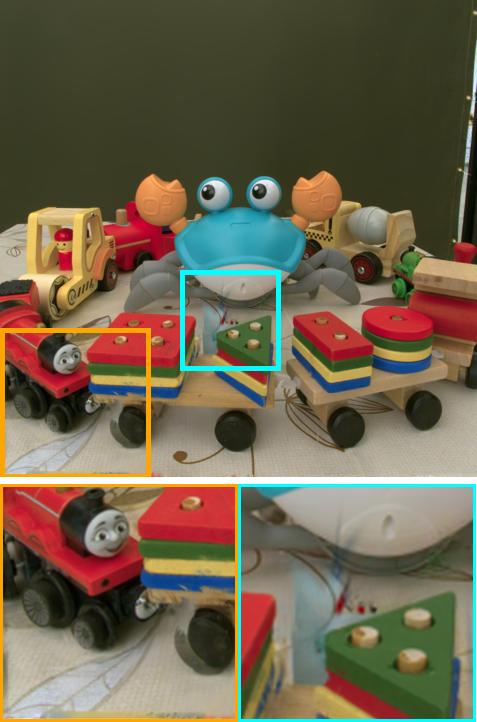}   
    \end{subfigure}    
    \hfill  
    \begin{subfigure}[t]{0.24\textwidth}
        \centering
        \includegraphics[width=\textwidth, keepaspectratio]{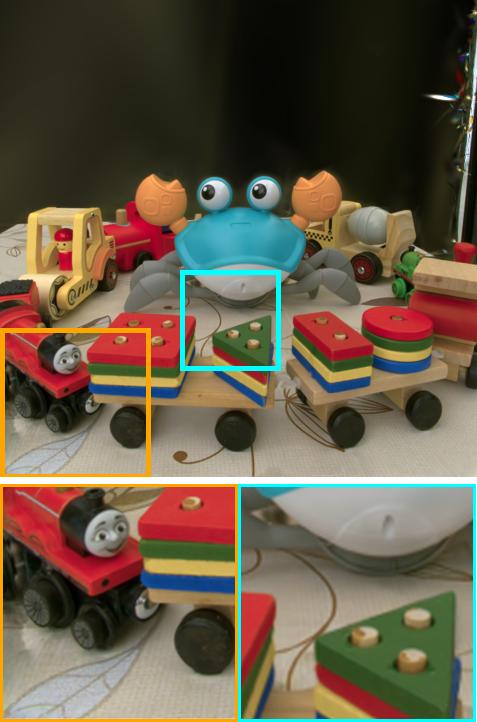}  
    \end{subfigure}    
    \hfill  
    \begin{subfigure}[t]{0.24\textwidth}
        \centering
        \includegraphics[width=\textwidth, keepaspectratio]{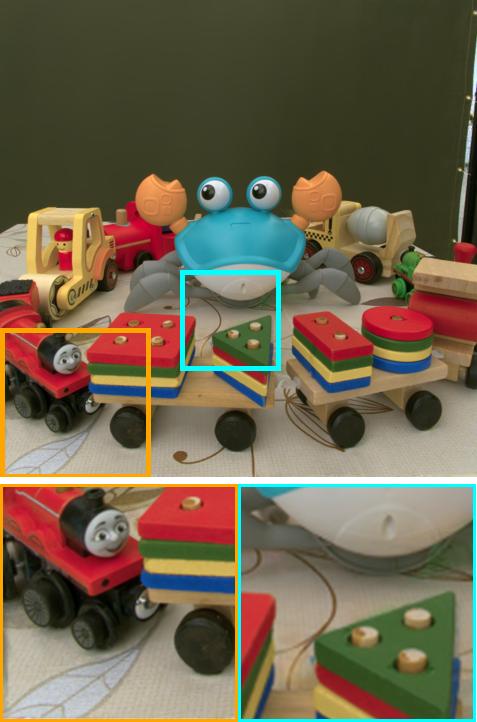}   
    \end{subfigure}    
    \hfill
    \begin{subfigure}[t]{0.24\textwidth}
        \centering
        \includegraphics[width=\textwidth, keepaspectratio]{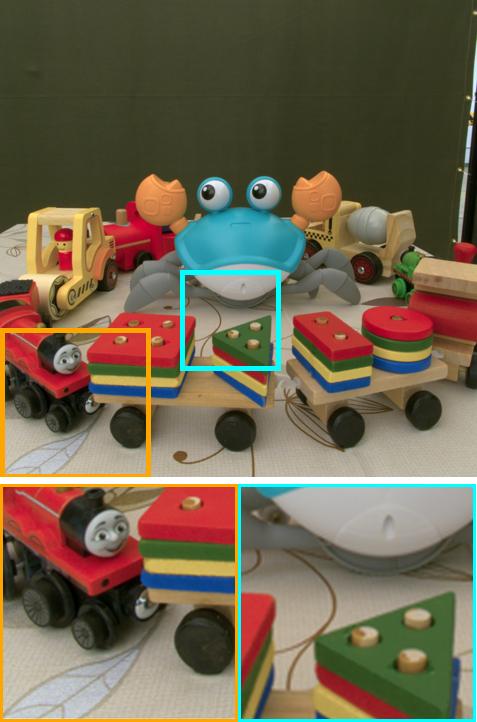} 
    \end{subfigure}

    \vspace{0.2cm}
 
    \begin{subfigure}[t]{0.24\textwidth}
        \centering
        \includegraphics[width=\textwidth, keepaspectratio]{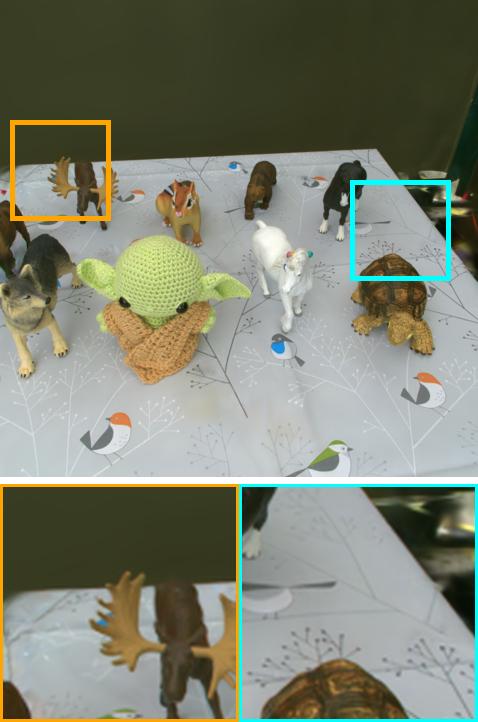}   
    \end{subfigure}    
    \hfill  
    \begin{subfigure}[t]{0.24\textwidth}
        \centering
        \includegraphics[width=\textwidth, keepaspectratio]{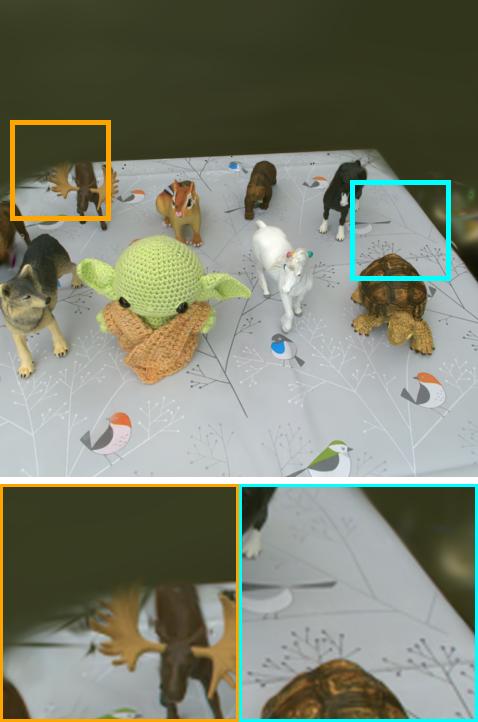}  
    \end{subfigure}    
    \hfill  
    \begin{subfigure}[t]{0.24\textwidth}
        \centering
        \includegraphics[width=\textwidth, keepaspectratio]{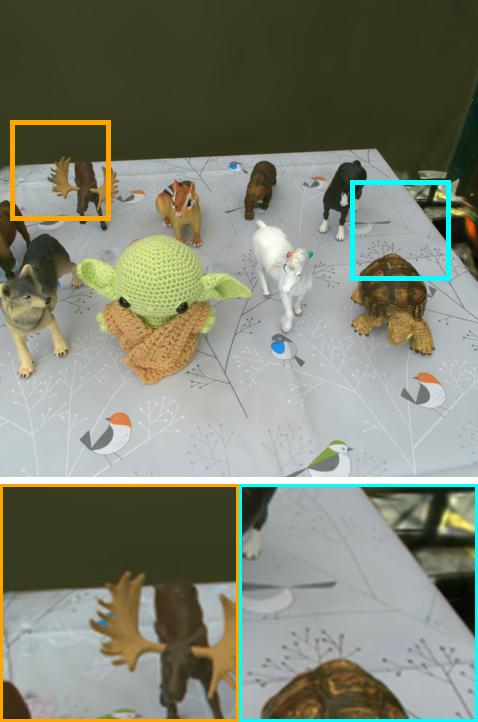}   
    \end{subfigure}    
    \hfill
    \begin{subfigure}[t]{0.24\textwidth}
        \centering
        \includegraphics[width=\textwidth, keepaspectratio]{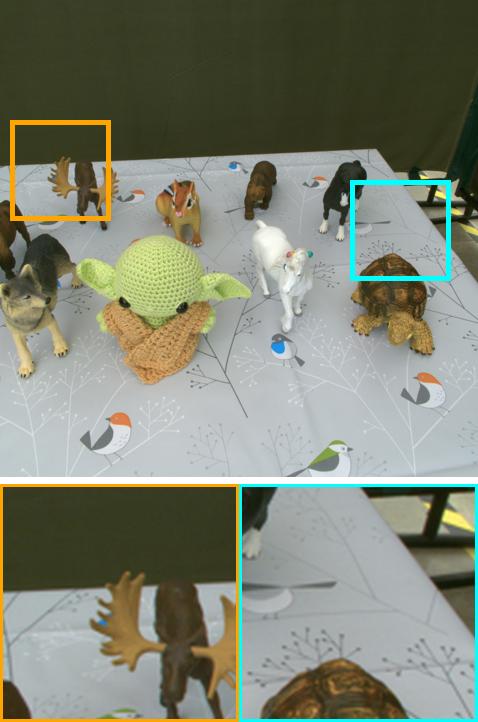} 
    \end{subfigure}

    \begin{subfigure}[t]{0.24\textwidth}
        \centering  
        w/o Segmentation
    \end{subfigure}    
    \hfill  
    \begin{subfigure}[t]{0.24\textwidth}
        \centering
        w/o Neural
    \end{subfigure}    
    \hfill  
    \begin{subfigure}[t]{0.24\textwidth}
        \centering  
        Robust Gaussian Splatting (\textbf{Ours})
    \end{subfigure}    
    \hfill
    \begin{subfigure}[t]{0.24\textwidth}
        \centering
        Ground truth 
    \end{subfigure}       
    
    \caption{Example comparison of all ablations. The background of w/o Neural is poor, but it filters the distractors efficiently. The w/o Segmentation version has a good background, but fails to remove all distractors and artifacts, resulting in blurred parts. We can see that our full version is most effective at ignoring distractors. Further ablation comparisons are provided in the supplementary material.}
    \vspace{-0.25cm}
    \label{fig:example_comparison}
\end{figure}

In \cref{fig:qual_2} we can see a baseline comparison of our method. As shown in the figure, 3D Gaussian Splatting captures the scene in high quality. However, it is filled to a large degree with artifacts of the distractors. When using RobustNeRF (GS), most of these distractors disappear. However, image quality significantly decreases, resulting in poor backgrounds and blurry subjects. When inspecting the masks generated by RobustNeRF (GS), it becomes clear that they are overly aggressive. While all distractors are filtered out, large parts of the remaining input image are masked out as well, effectively reducing the amount of training data. Our method performs best and reduces artifacts to a minimum while maintaining good backgrounds and an overall high image quality.

\Cref{fig:example_comparison} shows an ablation study of our method. 
The version w/o Segmentation achieves good reconstruction quality in presence of distractors, however, some artifacts and blurry parts remain, compared to the full version of our method. The version w/o Neural removes many distractor artifacts but struggles with the background and is sensitive to the segmentation quality of SegmentAnything. Robust Gaussian Splatting (ours) yields the best results, achieving good image quality in the foreground and background and only containing a minimum amount of artifacts. 
Further qualitative comparisons are provided in the supplementary material. We found that our method performs well in various scenarios and show a scene where a person acts as distractor in the supplementary material.

Overall, RobustNeRF (GS) is too aggressive in masking and thus decreases image quality. By refining the masking process, we ensure that our method maintains its effectiveness in handling occlusions and complex scenes while preserving the fidelity of reconstructed images. Our improvements make the masks less aggressive without adding artifacts. 

In \cref{tab:quant_results_clean} we show comparisons of our approach (Robust Gaussian Splatting) and 3D Gaussian Splatting \cite{kerbl20233d} on clean image datasets without distractors. We can see that our model performs fairly comparable to the original Gaussian Splatting on distractor-free scenes. Checking the distractor masks in the clean settings shows that they are almost all blank. This demonstrates that our approach rarely misclassifies static scene content as distractor. Slightly reduced metrics of the robust method in the clean setting is known topic, that was equally reported in RobustNeRF~\cite{sabour2023robustnerf}. Furthermore, Robust Gaussian Splatting is reliable in scenes with different amounts of distractors. For further detailed analysis see the supplementary material.
\begin{table*}[!h]
    \vspace{-0.5cm}
    \centering
    \caption{Comparison between our approach and the original 3D Gaussian Splatting on clean image datasets. The scene Playroom is from the Deep Blending Dataset \cite{hedman2018deep}, Truck from Tanks\&Temples \cite{knapitsch2017tanks} and Yoda (clean) from \cite{sabour2023robustnerf}.}
    \begin{tabularx}{\textwidth}{|>{\centering\arraybackslash}X|>{\centering\arraybackslash}X||>{\centering\arraybackslash}X|>{\centering\arraybackslash}X|>{\centering\arraybackslash}X|>{\centering\arraybackslash}X|>{\centering\arraybackslash}X|}
        \hline
        Model&Metric&Playroom&Truck&Yoda (clean)&Mean\\
        \hhline{|=|=#=|=|=|=|=|}
        \hline
        \multirow{3}{*}{\shortstack{Gaussian \\ Splatting}}&SSIM \(\uparrow\)&\textbf{0.9219}&\textbf{0.9244}&0.9277&\textbf{0.9247}\\
        &PSNR \(\uparrow\)& \textbf{30.34}&\textbf{26.82}&34.17&\textbf{30.43}\\
        &LPIPS \(\downarrow\)& \textbf{0.1403}&0.0787&0.1596&0.1262\\
        \hline
        \multirow{3}{*}{\shortstack{Robust \\ Gaussian \\ Splatting}}&SSIM&0.9174&0.9213&\textbf{0.9366}&0.8962\\
        &PSNR&30.05&25.71&\textbf{34.66}&30.14\\
        &LPIPS&0.1436&\textbf{0.0727}&\textbf{0.1432}&\textbf{0.11198}\\
        \hline    
    \end{tabularx}
    \vspace{-0.4cm}
    \label{tab:quant_results_clean}
\end{table*}

\subsection{Limitations}
\begin{figure}[!h]
    \centering
    \includegraphics[scale=0.6]{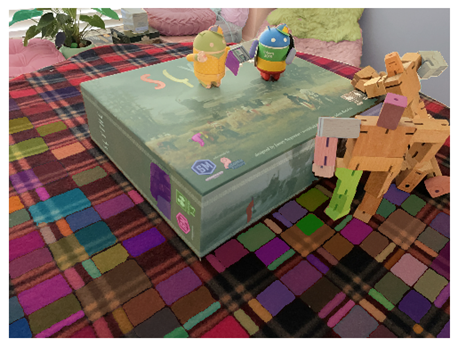}
    \caption{SegmentAnything struggles with correct object segmentation in some scenes. In this scene, each tile of the tablecloth is segmented as an object. This leads to worse distractor masks and therefore worse results.}
    \label{fig:limitations_sam}
\end{figure}

Our method reliably filters distractors and is able to render high quality novel views in scenes with different distractors. However, in some scenes SegmentAnything struggles to segment the correct objects. In \cref{fig:limitations_sam} we can see that SegmentAnything segments each tile of the tablecloth as an object. Using different hyperparameters for SegmentAnything in different scenes can solve this problem and can be addressed in future research. Nevertheless, as we can see in the detailed per scene evaluations in the supplementary material our method performs well considering the incorrect segmentation of SAM. 

Furthermore, it is possible to tune the regularization strength and the minimum intersection constant from \cref{eq:intersections} for individual scenes to get even better performance. 
However, we use the same parameter set across all tested scenes and found it generalizes well.

\section{Conclusion}
We address the problem of novel view synthesis in the presence of distractors using 3D Gaussian Splatting. We were consistently able to generate high-quality novel views from input data polluted by distractors. By introducing learnable neural decision boundaries and object awareness into the distractor tracking, we achieve considerably better quantitative and qualitative results than the RobustNeRF (GS and NeRF) method and 3D Gaussian Splatting. 

In addition, Robust Gaussian Splatting maintains fairly comparable performance to Gaussian Splatting on clean scenes without distractors. 
Overall, we believe that our approach is an important step towards robust high-quality novel view synthesis that is easily applicable to in-the-wild data, where the efforts and costs of scene capture and data preprocessing can be kept minimal.

% Note: if you want to use up all of the allowed space for the paper,
%       the bibliography will start on top of page 13. Furthermore,
%       from page 13 onwards, there will be *only* bibliography, no more
%       figures/tables.
\bibliographystyle{splncs04}
\bibliography{main}

\newpage
\appendix

\section{Training Details}
\label{sec:training_details}
We use a logistic regression as the neural decision boundary trained using SGD with a learning rate of $0.1$. For the mask loss we have a regularization strength of $0.1$. Furthermore, we use an intersection ratio of 40\%. Other 3D Gaussian Splatting hyperparameters are set to default. All experiments are trained for 30k iterations. All scenes use the same hyperparameters.

The models on clean scenes were trained for 30k iterations as well. Images from the scene Playroom and Truck are downsized by the factor of 2 and Yoda (clean) is downsized by 8.

\section{Analysis of Fraction of Cluttered Images inside a Scene}
\vspace{-0.8cm}
\label{sec:cluttered_analysis}
\begin{figure}[!h]
    \centering
    \scalebox{1.}{%
    % \begin{tikzpicture}[scale=1]
    % \begin{axis}[
    %     title={},
    %     xlabel={Fraction of images containing distractor},
    %     ylabel={LPIPS},
    %     xmin=0, xmax=100,
    %     ymin=0.1, ymax=0.2,
    %     legend pos= outer north east,
    %     ymajorgrids=true,
    %     grid style=dashed,
    % ]
    
    % \addplot[
    %     color=blue,
    %     mark=square,
    %     ]
    %     coordinates {
    %     (0,0.1596246)(20,0.1477044)(40,0.1518471)(60,0.1707504)(80,0.1687387)(100,0.1820882)
    %     };
    %     \addlegendentry{Gaussian Splatting}
    
    % \addplot[
    %     color=red,
    %     mark=square,
    %     ]
    %     coordinates {
    %     (0,0.1432159)(20,0.1394680)(40,0.1364592)(60,0.1447061)(80,0.1504304)(100,0.1604359)
    %     };
    %     \addlegendentry{Robust Gaussian Splatting}
    % \end{axis}
    % \end{tikzpicture}
    \includegraphics[]{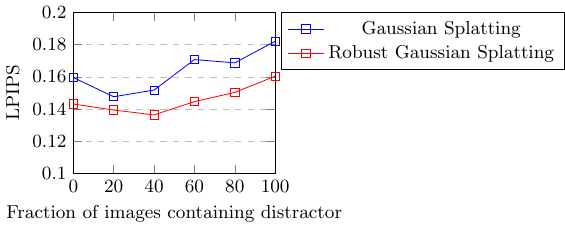}
    }
    \caption{Metric for the Yoda scene with different fractions of cluttered images. Our method performs better than Gaussian Splatting on all percentages of cluttered images. }
    \label{fig:cluttered-clean}
\end{figure}
Robust Gaussian Splatting is very reliable in scenes with different amounts of distractors. \cref{fig:cluttered-clean} presents a comparative analysis of our Robust Gaussian Splatting technique against 3D Gaussian Splatting in the Yoda scene under differing proportions of cluttered and clean images. A cluttered image is an image containing distractors. In particular, as the fraction of cluttered images increases, Gaussian Splatting struggles to maintain accuracy. Its performance declines, dropping from 0.159 on clean images to 0.182 on images containing only distractors. In contrast, our method exhibits remarkable stability, maintaining accuracy levels around 0.14-0.16. This robustness to the percentage of cluttered images highlights the effectiveness of our approach, particularly in challenging scenarios characterized by high levels of clutter.

\vspace{-0.2cm}
\section{Detailed Quantitative Analysis}
\vspace{-0.1cm}
\label{sec:detailed_quant_analysis}
\cref{tab:quant_results_appendix} shows that the w/o segmentation method yields less consistent improvements. This could be traced back to the high variance in the accuracy of the segments provided by SegmentAnything. For example, in the and-bot scene, each square of the checkered tablecloth at its center is a separate segment, while the distractors are often not part of any segment. This could be prevented by using different hyperparameters for SegmentAnything, but those might not work for other scenes. The segmentation still considerably improves scores when applied on top of the neural decision boundary optimization, yielding improved scores for our full method over the w/o Segmentation ablation leading to the best mean improvement. Overall, our improvements perform considerably better.

\begin{table*}[!h]
    \centering
        \caption{Quantitative comparison to baselines and ablated versions. The full version of our method performs better than Gaussian Splatting and RobustNeRF, as well as the ablations of our method. Values for the RobustNeRF (NeRF) implementation are taken from \cite{sabour2023robustnerf}.}
    \begin{tabularx}{\textwidth}{|>{\centering\arraybackslash}X|>{\centering\arraybackslash}X||>{\centering\arraybackslash}X|>{\centering\arraybackslash}X|>{\centering\arraybackslash}X|>{\centering\arraybackslash}X|>{\centering\arraybackslash}X|}
        \hline
        Model&Metric&Statue&Yoda&And-bot&Crab&Mean\\
        \hhline{|=|=#=|=|=|=|=|}
        \hline
        \multirow{3}{*}{\shortstack{Gaussian \\ Splatting}}&SSIM \(\uparrow\)&0.8401&0.9111&0.8004&0.9385&0.87253\\
        &PSNR \(\uparrow\)&21.56&29.99&23.63&30.92&26.53\\
        &LPIPS \(\downarrow\)&0.1443&0.1821&0.15938&0.1414&0.1568\\
        \hline
        \multirow{3}{*}{\shortstack{Robust \\ NeRF \\ (GS)}}&SSIM&0.8410&0.8821&0.7952&0.9132&0.8579\\
        &PSNR&21.27&25.93&22.71&26.46&24.09\\
        &LPIPS&0.2001&0.2177&0.1746&0.1891&0.1773\\
        \hline
        \multirow{3}{*}{\shortstack{Robust \\ NeRF \\ (NeRF)}}&SSIM& 0.75 & 0.83 &  0.65 & 0.81 & 0.76\\
        &PSNR& 20.89 & 30.87 & 21.72 & 30.75 & 26.06\\
        &LPIPS& 0.28 & 0.20 & 0.31 & 0.21 & 0.25\\
        \hline
        \multirow{3}{*}{\shortstack{w/o \\ Segment- \\ation}}&SSIM&0.84164&\textbf{0.9236}&0.8235&0.9481&0.8842\\
        &PSNR&21.52&32.41&24.40&33.46&27.95\\
        &LPIPS&0.1376&\textbf{0.1584}&\textbf{0.1308}&0.1184&0.1363\\
        \hline
        \multirow{3}{*}{\shortstack{ w/o Neural}}&SSIM&0.8268&0.8994&0.7794&0.8034&0.8273\\
         &PSNR&20.21&29.76&21.51&21.62&23.28\\
        &LPIPS&0.1593&0.1654&0.1778&0.1765&0.1698\\
        \hline
        \multirow{3}{*}{\shortstack{Robust \\ Gaussian \\ Splatting}}& SSIM&\textbf{0.8514}&0.9235&\textbf{0.8240}&\textbf{0.9511}&\textbf{0.8875}\\
        &PSNR&\textbf{22.21}&\textbf{32.65}&\textbf{24.48}&\textbf{34.23}&\textbf{28.39}\\
        &LPIPS&\textbf{0.1214}&0.1604&0.1314&\textbf{0.1151}&\textbf{0.1321}\\
        \hline
    \end{tabularx}
    \label{tab:quant_results_appendix}
\end{table*}

\newpage
\section{Further Qualitative Comparisons}
\label{sec:further_qual_comp}

\begin{figure}[!hp]
    \centering  
    \begin{subfigure}[t]{0.24\textwidth}
        \centering
        \includegraphics[width=\textwidth, keepaspectratio]{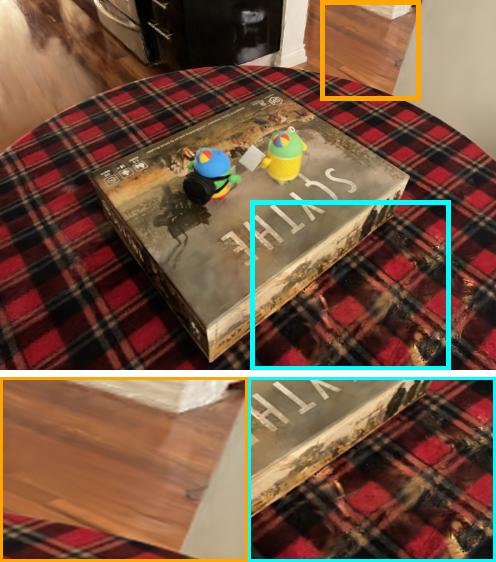}   
    \end{subfigure}    
    \hfill  
    \begin{subfigure}[t]{0.24\textwidth}
        \centering
        \includegraphics[width=\textwidth, keepaspectratio]{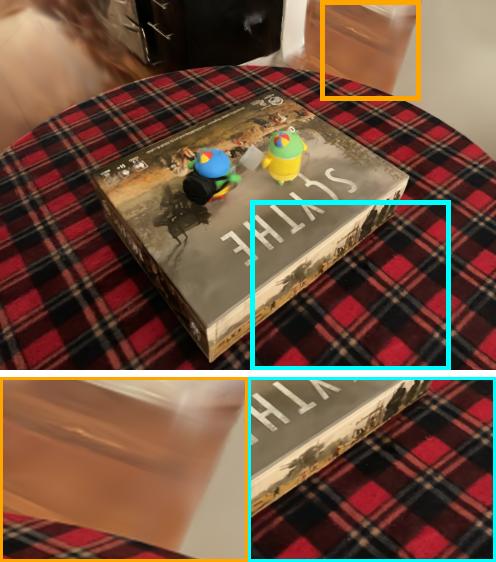}   
    \end{subfigure}    
    \hfill    
    \hfill  
    \begin{subfigure}[t]{0.24\textwidth}
        \centering
        \includegraphics[width=\textwidth, keepaspectratio]{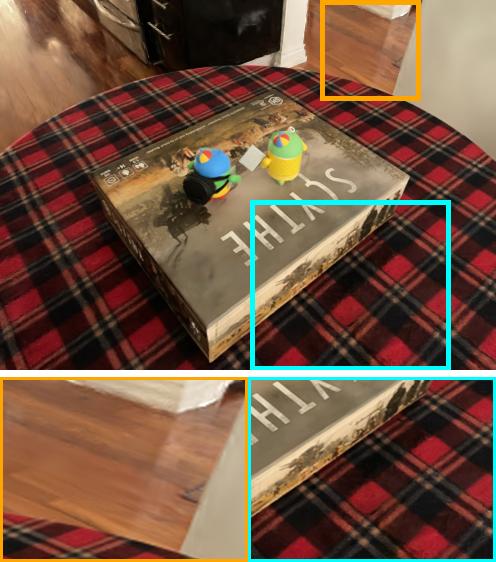}   
    \end{subfigure}    
    \hfill
    \begin{subfigure}[t]{0.24\textwidth}
        \centering
        \includegraphics[width=\textwidth, keepaspectratio]{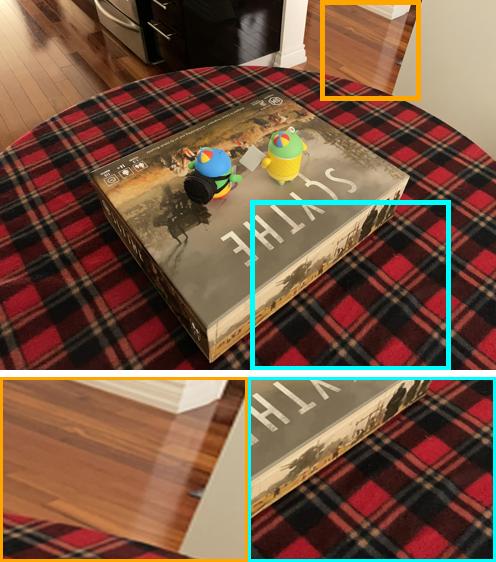} 
    \end{subfigure}

    \vspace{0.2cm}

    \begin{subfigure}[t]{0.24\textwidth}
        \centering
        \includegraphics[width=\textwidth, keepaspectratio]{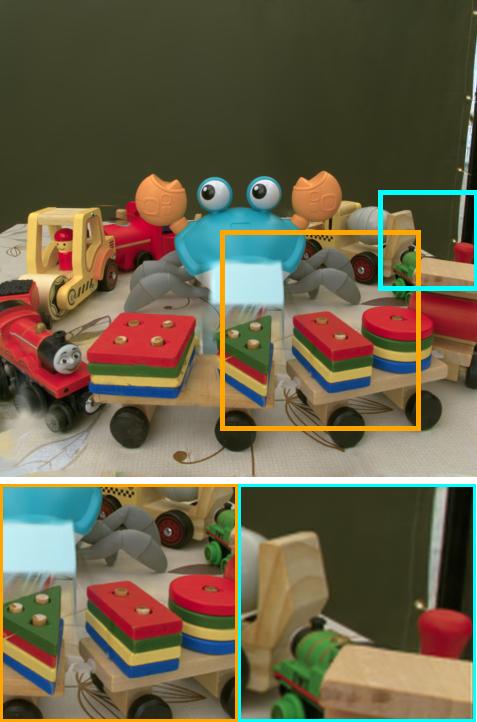}   
    \end{subfigure}    
    \hfill  
    \begin{subfigure}[t]{0.24\textwidth}
        \centering
        \includegraphics[width=\textwidth, keepaspectratio]{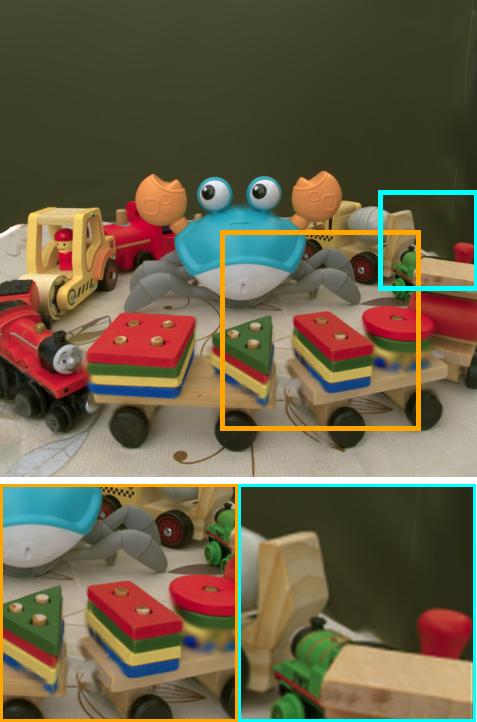}   
    \end{subfigure}      
    \hfill  
    \begin{subfigure}[t]{0.24\textwidth}
        \centering
        \includegraphics[width=\textwidth, keepaspectratio]{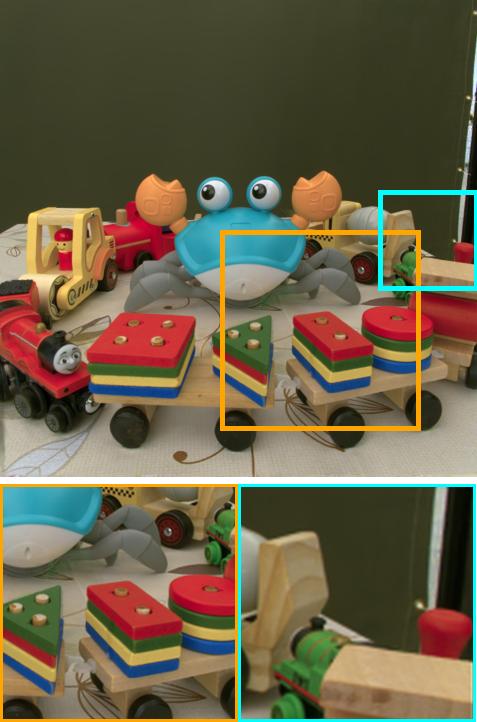}   
    \end{subfigure}    
    \hfill
    \begin{subfigure}[t]{0.24\textwidth}
        \centering
        \includegraphics[width=\textwidth, keepaspectratio]{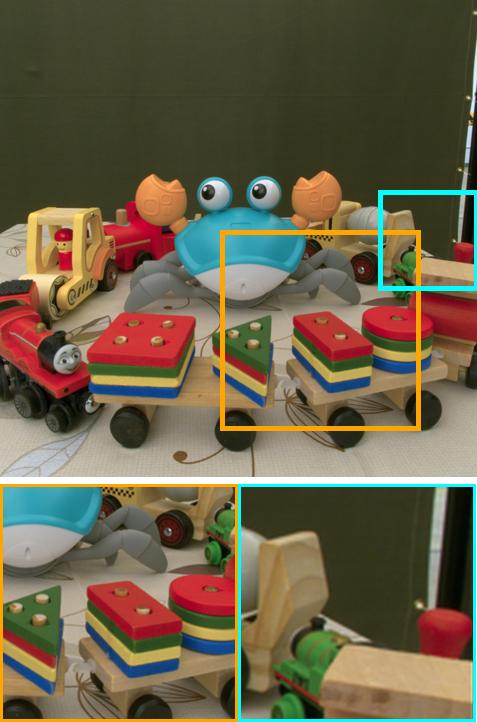} 
    \end{subfigure}
        
    \vspace{0.2cm}

    \begin{subfigure}[t]{0.24\textwidth}
        \centering
        \includegraphics[width=\textwidth, keepaspectratio]{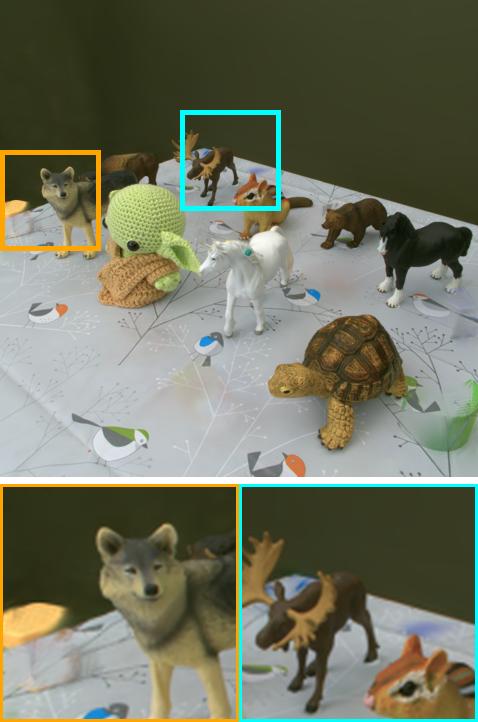}   
    \end{subfigure}    
    \hfill  
    \begin{subfigure}[t]{0.24\textwidth}
        \centering
        \includegraphics[width=\textwidth, keepaspectratio]{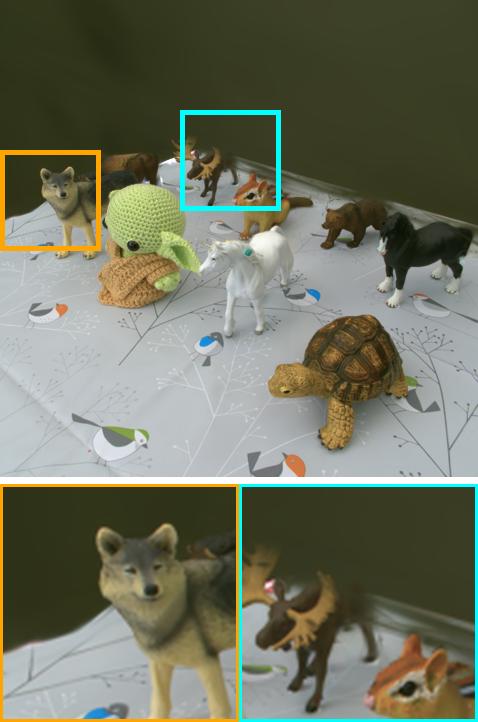}   
    \end{subfigure}    
    \hfill   
    \begin{subfigure}[t]{0.24\textwidth}
        \centering
        \includegraphics[width=\textwidth, keepaspectratio]{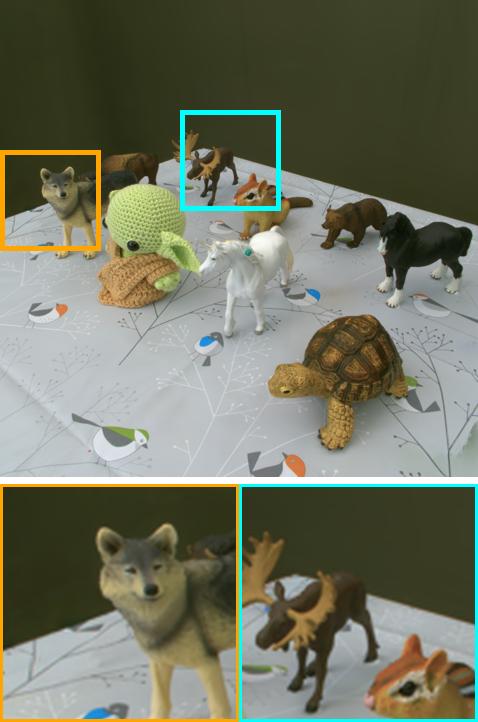}  
    \end{subfigure}    
    \hfill
    \begin{subfigure}[t]{0.24\textwidth}
        \centering
        \includegraphics[width=\textwidth, keepaspectratio]{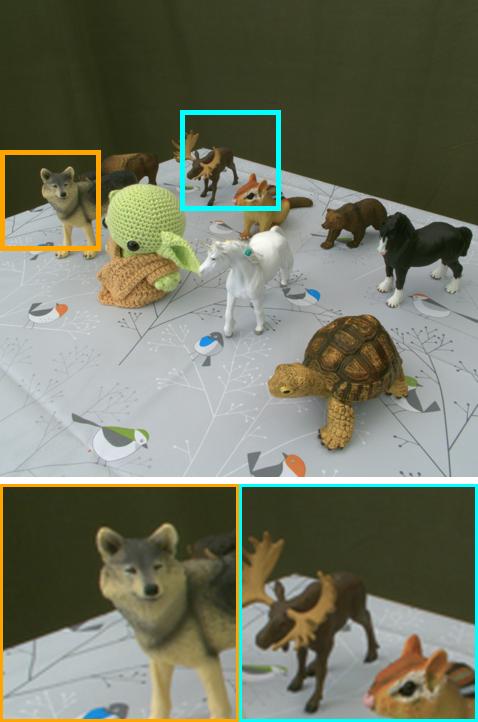} 
    \end{subfigure}

    \begin{subfigure}[t]{0.24\textwidth}
        \centering
        Gaussian Splatting
    \end{subfigure}    
    \hfill  
    \begin{subfigure}[t]{0.24\textwidth}
        \centering
        RobustNeRF (GS)
    \end{subfigure}    
    \hfill   
    \begin{subfigure}[t]{0.24\textwidth}
        \centering  
        Robust Gaussian Splatting (\textbf{Ours})
    \end{subfigure}    
    \hfill
    \begin{subfigure}[t]{0.24\textwidth}
        \centering
        Ground truth 
    \end{subfigure}       
    
    \caption{Example comparison of qualitative results from held-out test views. We can see that our full version is most effective in ignoring distractors.}
    \label{fig:qual_2_appendix}
\end{figure}

\newpage

\begin{figure}[!hp]
    \centering    
    \begin{subfigure}[t]{0.24\textwidth}
        \centering
        \includegraphics[width=\textwidth, keepaspectratio]{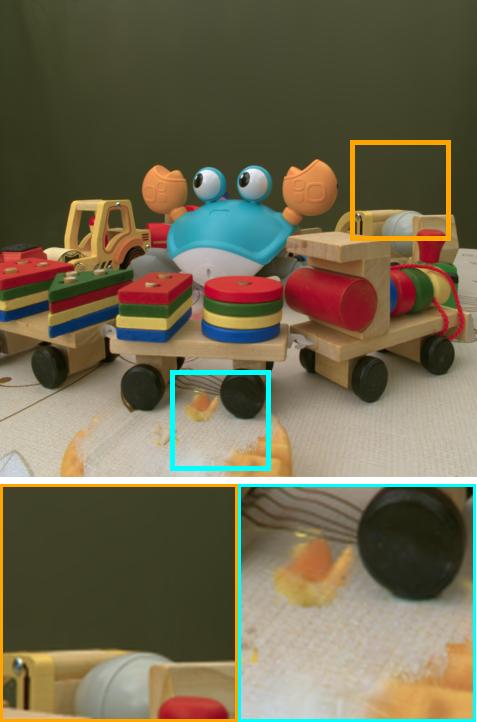}   
    \end{subfigure}    
    \hfill  
    \begin{subfigure}[t]{0.24\textwidth}
        \centering
        \includegraphics[width=\textwidth, keepaspectratio]{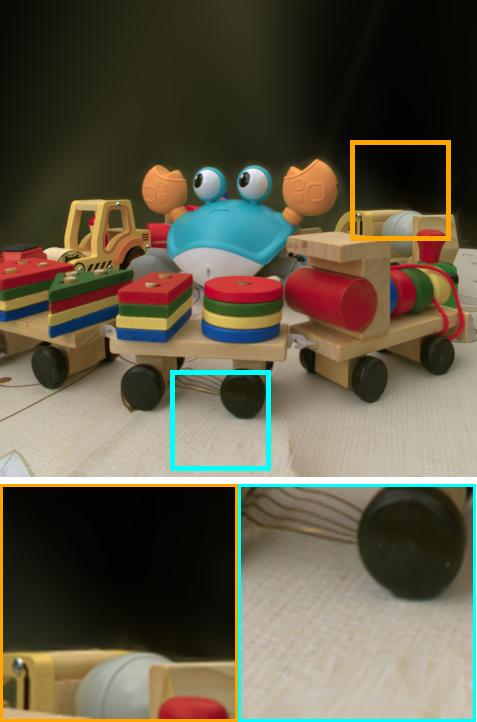}  
    \end{subfigure}    
    \hfill  
    \begin{subfigure}[t]{0.24\textwidth}
        \centering
        \includegraphics[width=\textwidth, keepaspectratio]{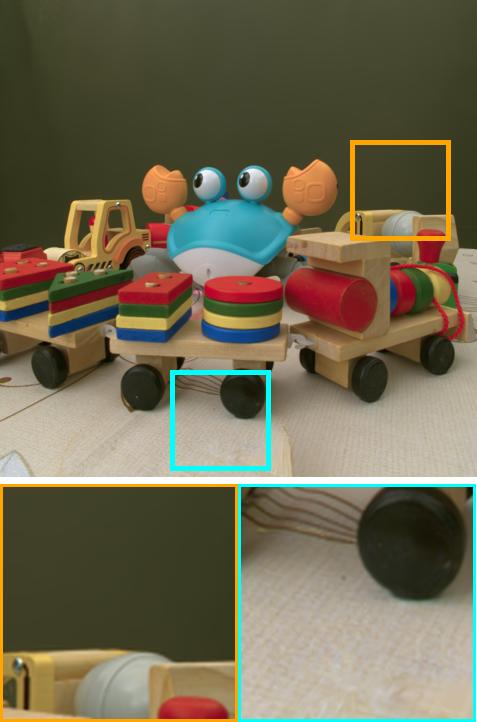}   
    \end{subfigure}    
    \hfill
    \begin{subfigure}[t]{0.24\textwidth}
        \centering
        \includegraphics[width=\textwidth, keepaspectratio]{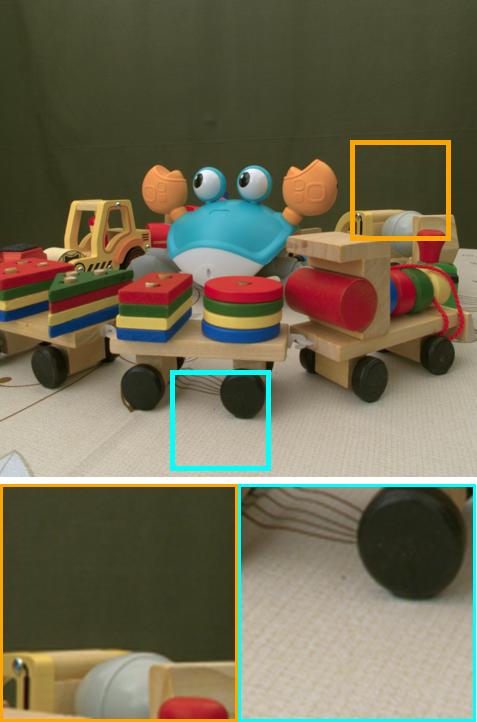} 
    \end{subfigure}

    \vspace{0.2cm}

    \begin{subfigure}[t]{0.24\textwidth}
        \centering
        \includegraphics[width=\textwidth, keepaspectratio]{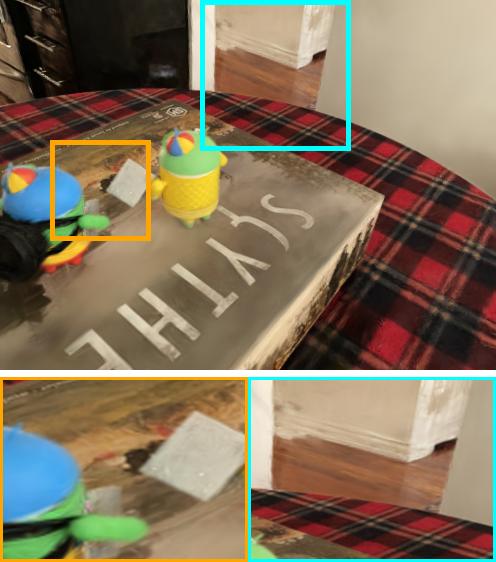}   
    \end{subfigure}    
    \hfill  
    \begin{subfigure}[t]{0.24\textwidth}
        \centering
        \includegraphics[width=\textwidth, keepaspectratio]{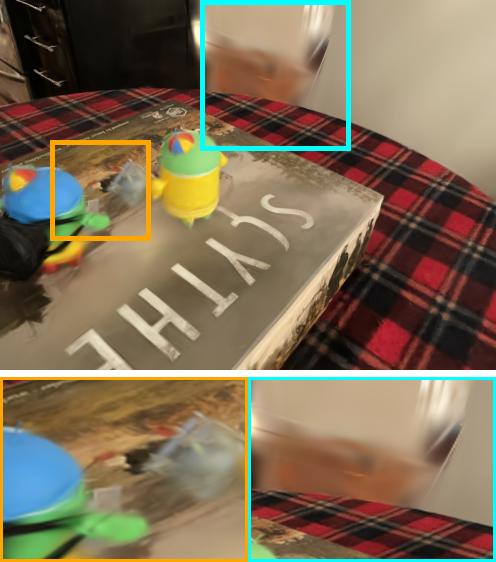}  
    \end{subfigure}    
    \hfill  
    \begin{subfigure}[t]{0.24\textwidth}
        \centering
        \includegraphics[width=\textwidth, keepaspectratio]{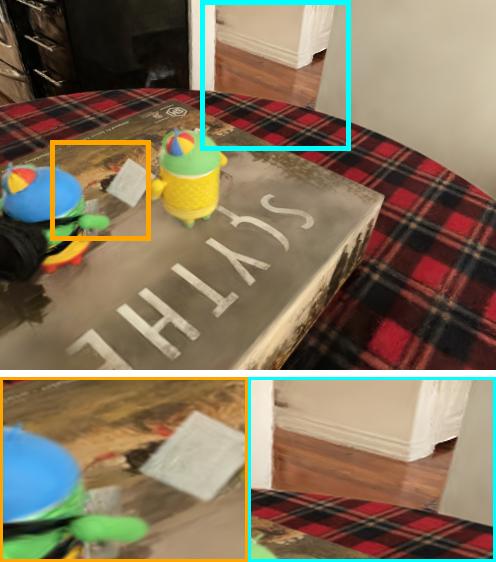}   
    \end{subfigure}    
    \hfill
    \begin{subfigure}[t]{0.24\textwidth}
        \centering
        \includegraphics[width=\textwidth, keepaspectratio]{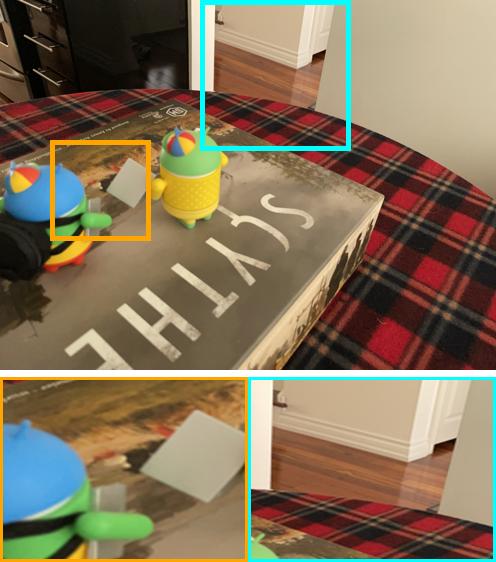} 
    \end{subfigure}
    
    \vspace{0.2cm}

    \begin{subfigure}[t]{0.24\textwidth}
        \centering
        \includegraphics[width=\textwidth, keepaspectratio]{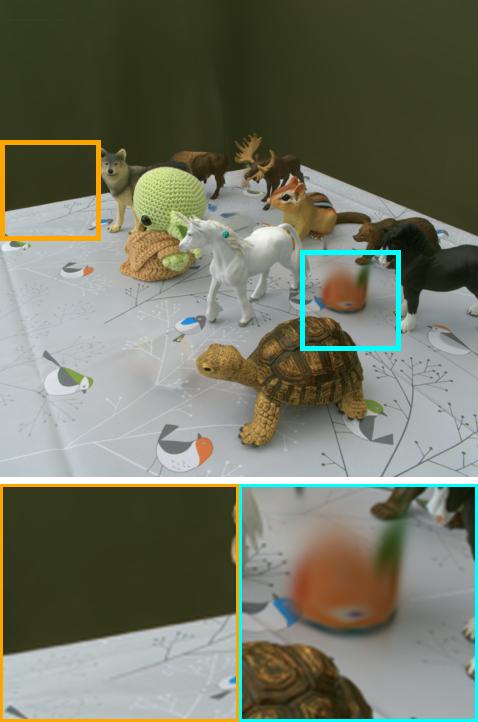}   
    \end{subfigure}    
    \hfill  
    \begin{subfigure}[t]{0.24\textwidth}
        \centering
        \includegraphics[width=\textwidth, keepaspectratio]{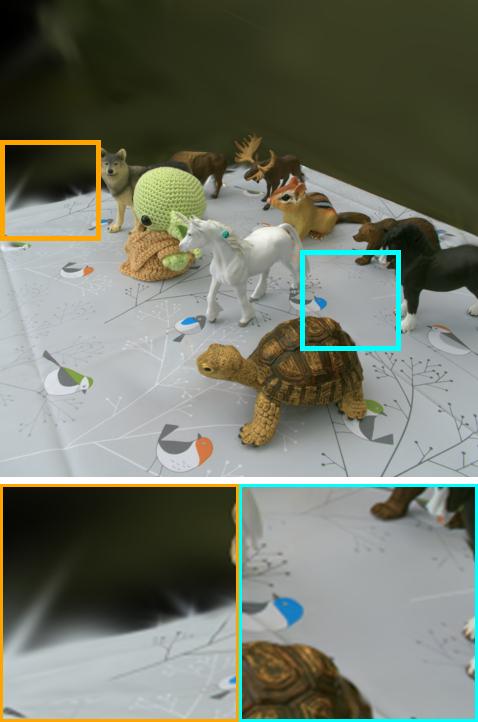}  
    \end{subfigure}    
    \hfill  
    \begin{subfigure}[t]{0.24\textwidth}
        \centering
        \includegraphics[width=\textwidth, keepaspectratio]{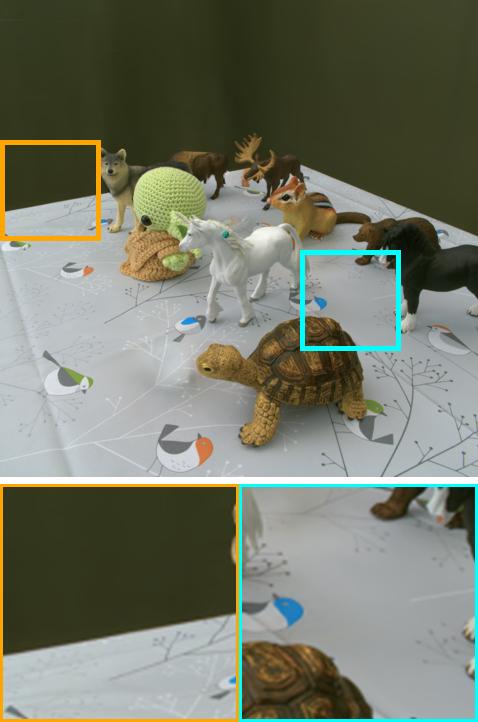}   
    \end{subfigure}    
    \hfill
    \begin{subfigure}[t]{0.24\textwidth}
        \centering
        \includegraphics[width=\textwidth, keepaspectratio]{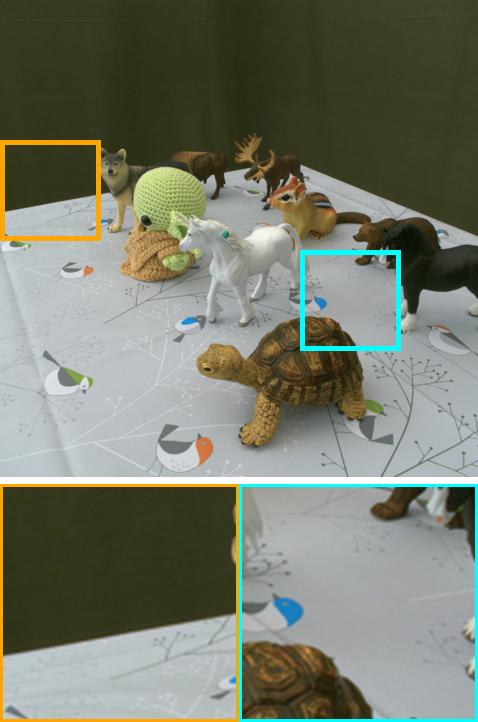} 
    \end{subfigure}

    \begin{subfigure}[t]{0.24\textwidth}
        \centering  
        w/o Segmentation
    \end{subfigure}    
    \hfill  
    \begin{subfigure}[t]{0.24\textwidth}
        \centering
        w/o Neural
    \end{subfigure}    
    \hfill  
    \begin{subfigure}[t]{0.24\textwidth}
        \centering  
        Robust Gaussian Splatting (\textbf{Ours})
    \end{subfigure}    
    \hfill
    \begin{subfigure}[t]{0.24\textwidth}
        \centering
        Ground truth 
    \end{subfigure}       
    
    \caption{Comparison of ablations using results from held-out test views. The w/o Neural version struggles to maintain good background but manages to minimize distractors. w/o Segmentation maintains good background but fails to filter out all distractors. We can see that our full version is most effective at ignoring distractors. }
    \label{fig:qual_2_appendix}
\end{figure}

\newpage

\section{Reallife Scene}
\label{sec:real_life}
\begin{figure}[!h]  
    \centering
    \begin{subfigure}[t]{0.32\textwidth}
         \centering        
         \begin{subfigure}[t]{0.48\textwidth}
             \centering
            \includegraphics[height=2.4cm]{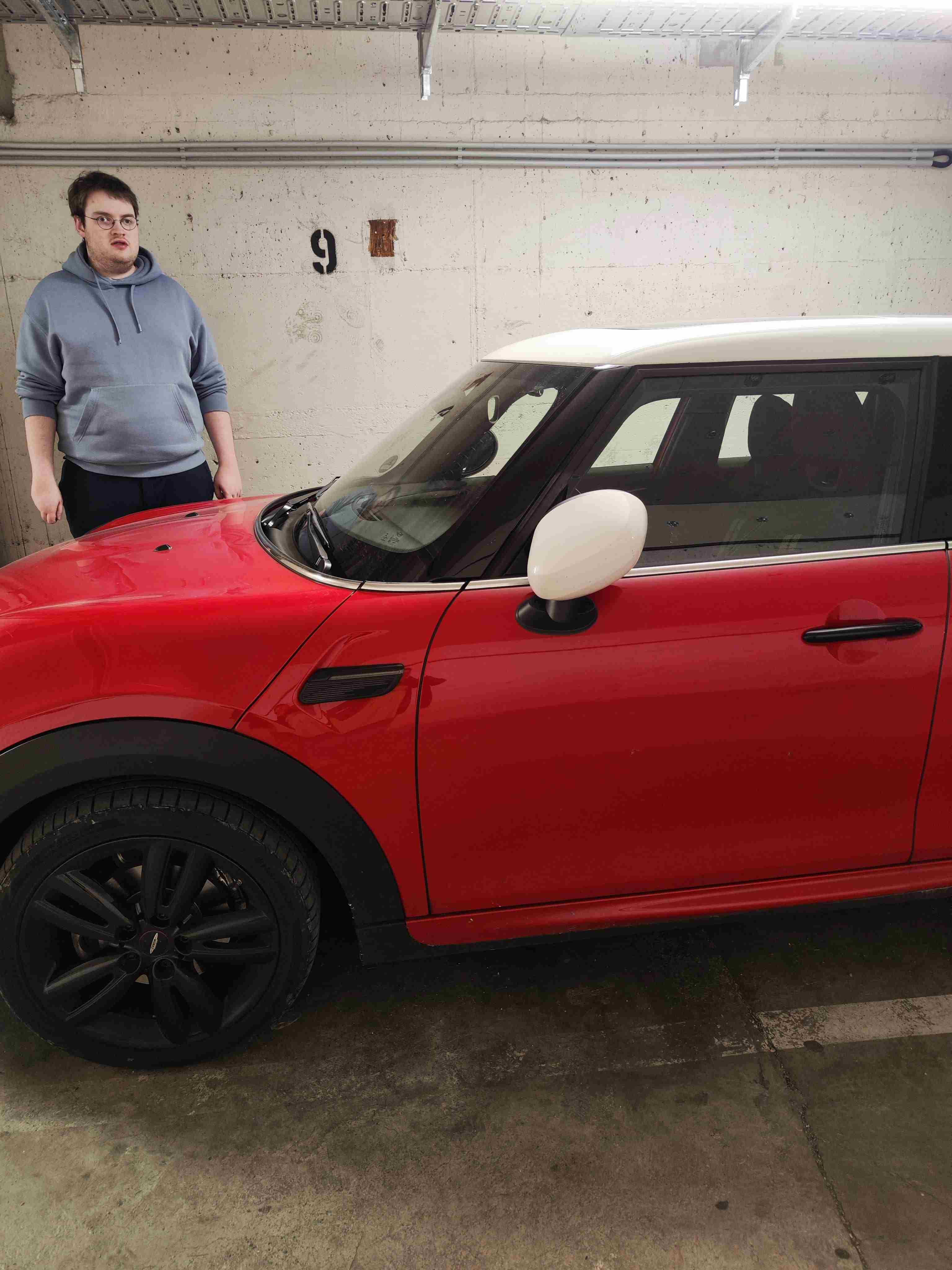}
         \end{subfigure}
         \begin{subfigure}[t]{0.48\textwidth}
             \centering
            \includegraphics[height=2.4cm]{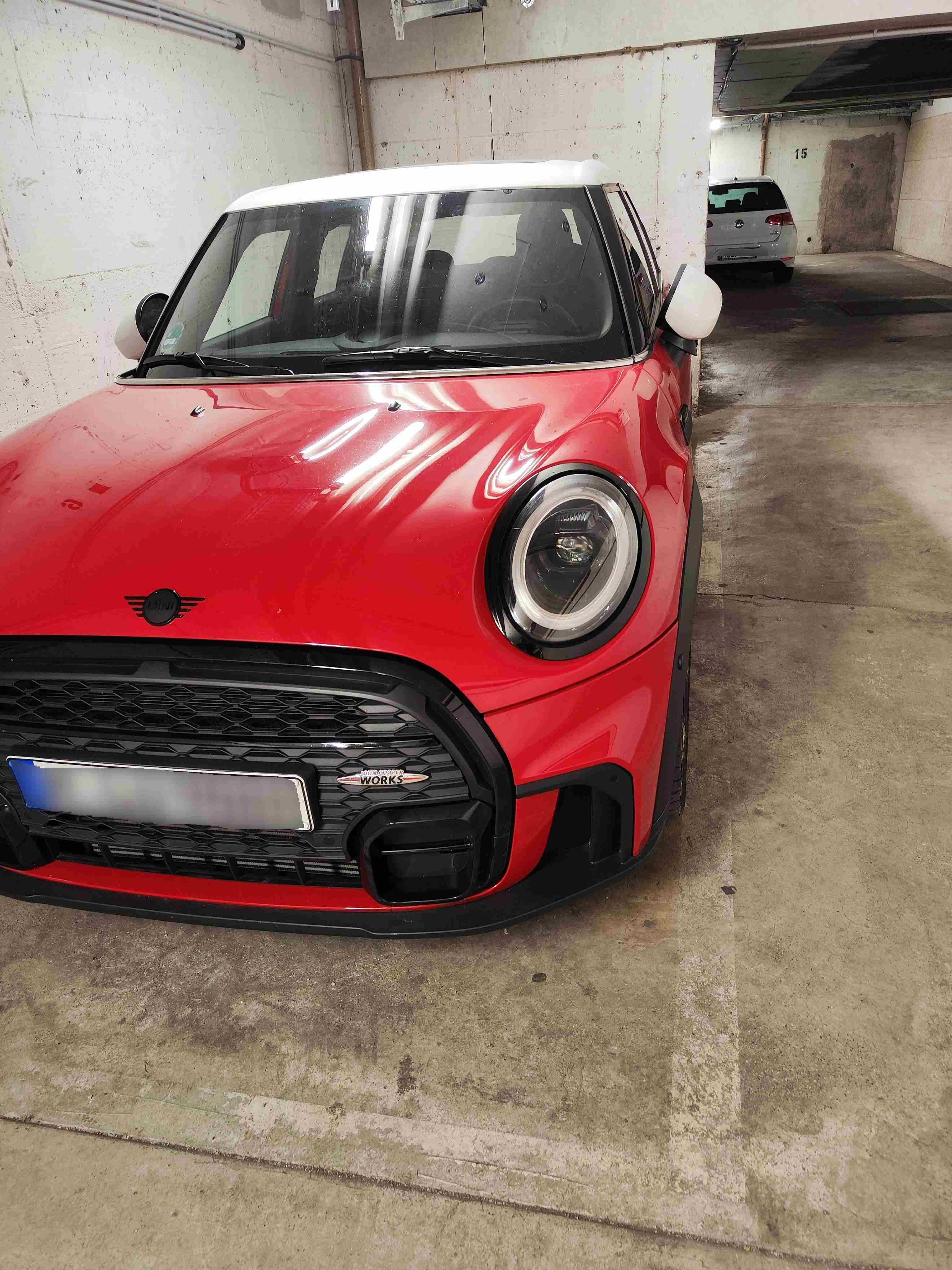}
         \end{subfigure}
        \caption{Example training images}
    \end{subfigure}
     \begin{subfigure}[t]{0.32\textwidth}
        \begin{subfigure}[t]{0.48\textwidth}
             \centering
            \includegraphics[height=2.4cm]{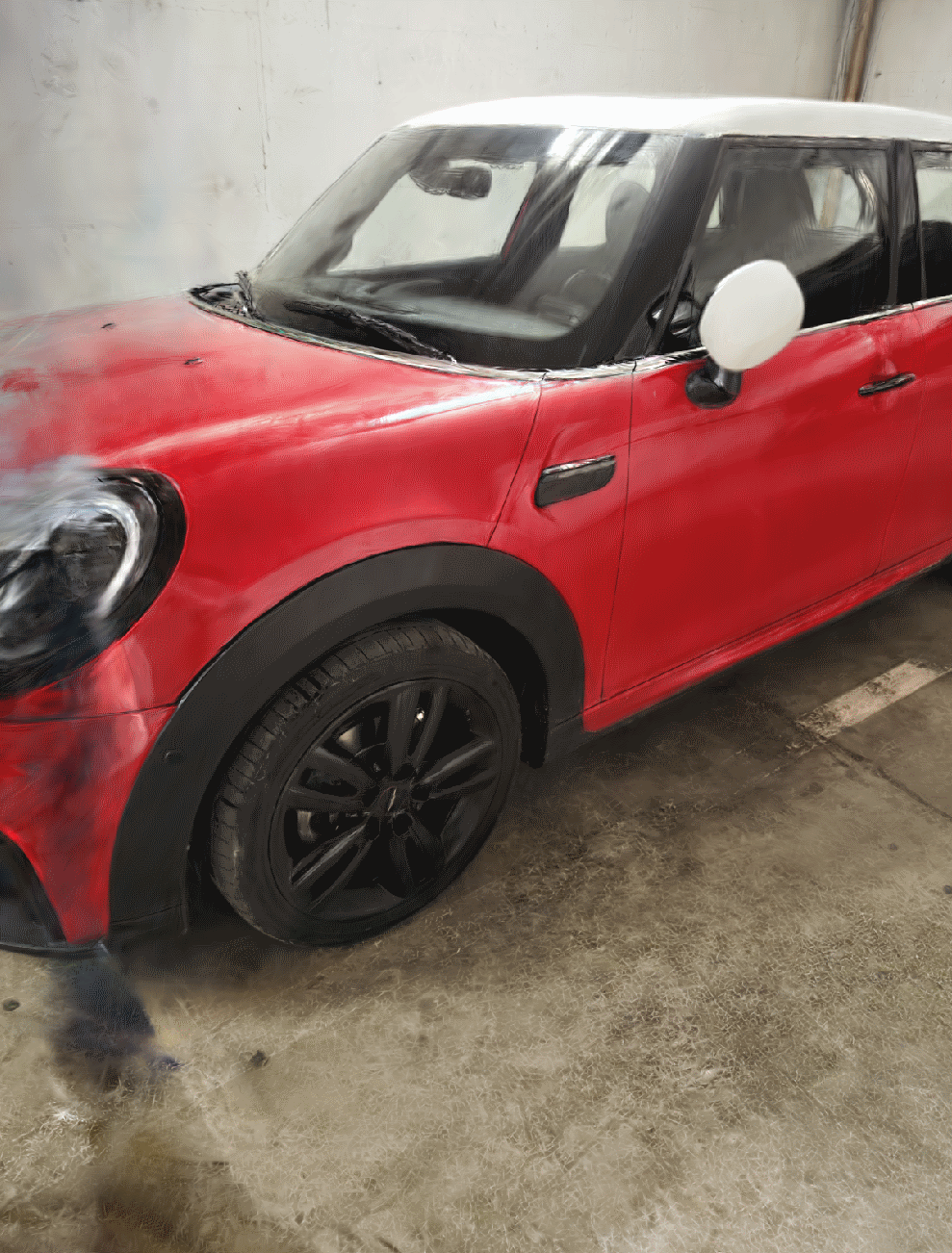}
        \end{subfigure}
        \begin{subfigure}[t]{0.48\textwidth}
             \centering
            \includegraphics[height=2.4cm]{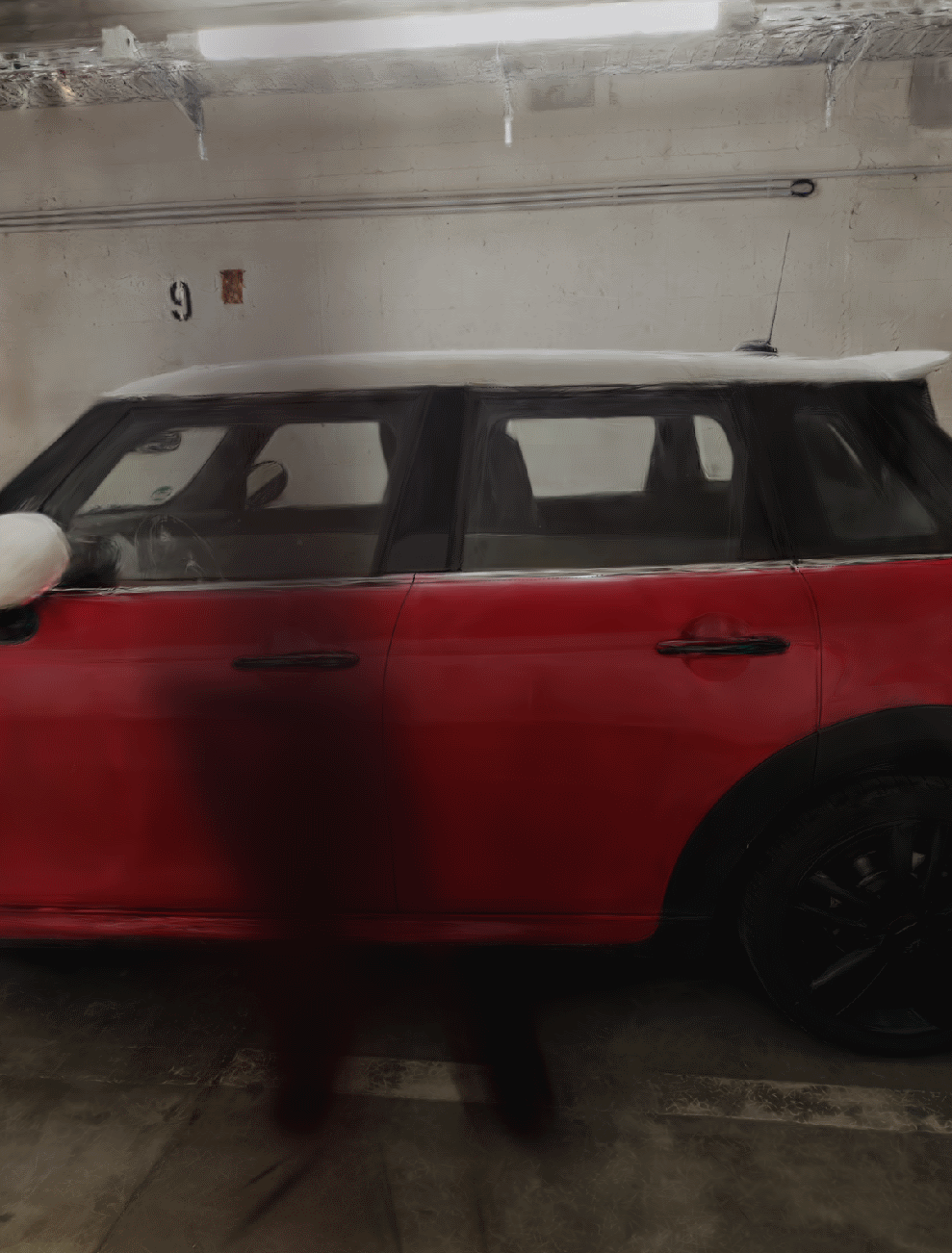}
        \end{subfigure}
        \caption{Gaussian Splatting}
     \end{subfigure}
    \begin{subfigure}[t]{0.32\textwidth}
        \begin{subfigure}[t]{0.48\textwidth}
             \centering
            \includegraphics[height=2.4cm]{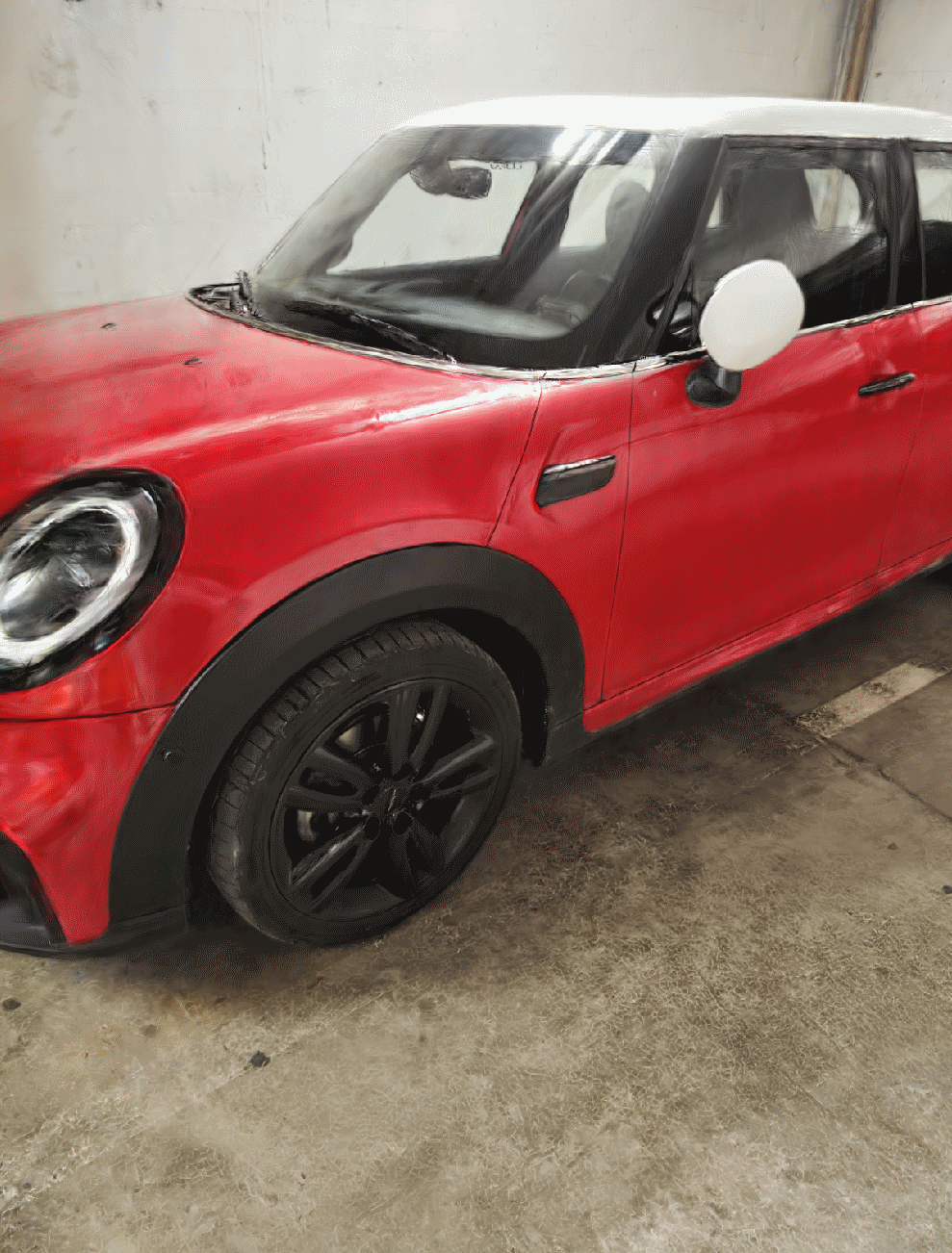}
        \end{subfigure}
        \begin{subfigure}[t]{0.48\textwidth}
             \centering
            \includegraphics[height=2.4cm]{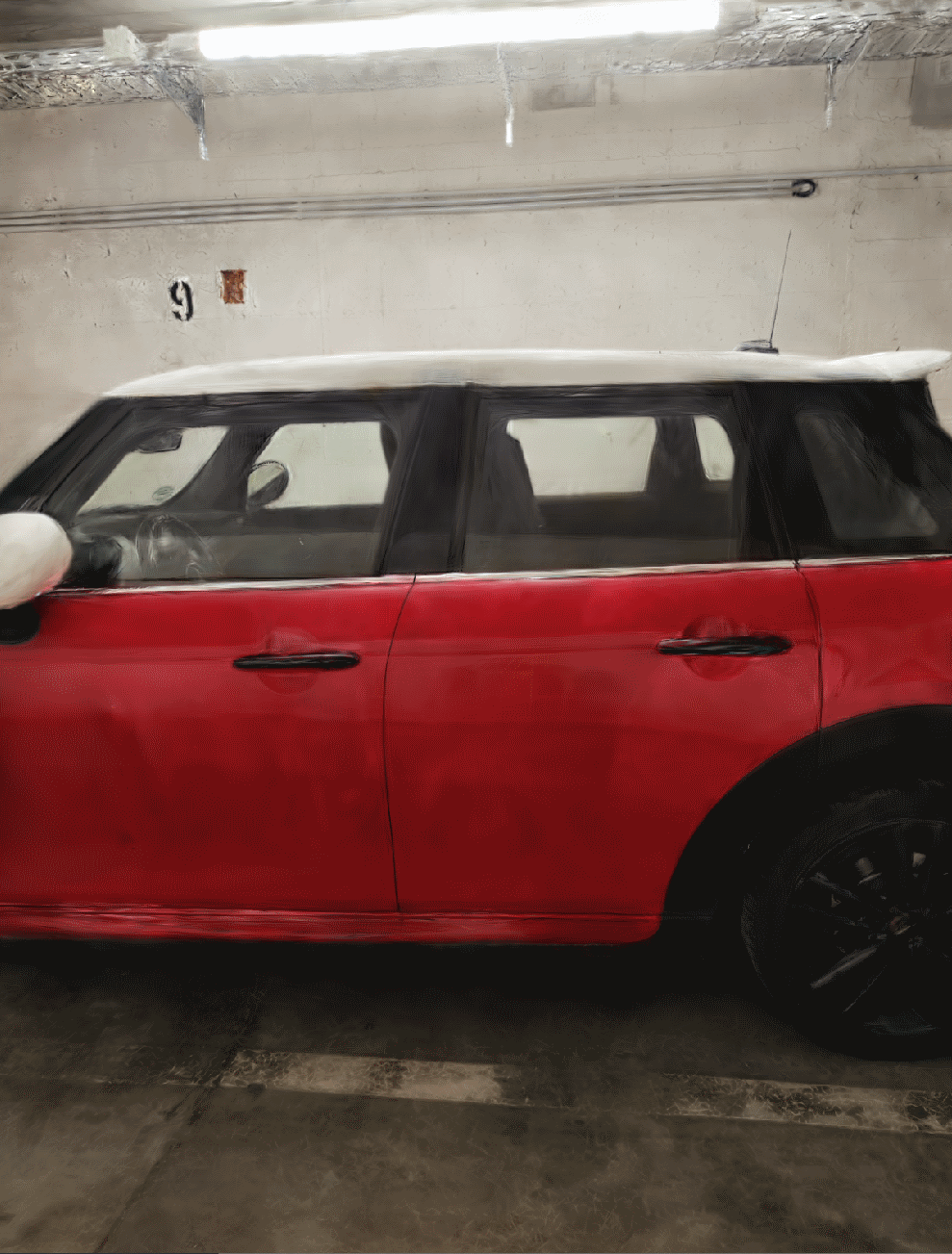}
        \end{subfigure}
        \caption{Robust Gaussian Splatting (\textbf{Ours})}
     \end{subfigure}
    \caption{Comparison of our method and Gaussian Splatting. We trained this scene with the same hyperparameters as from the main paper. 21\% of the images contain distractors. Our method effectively filters out distractors and provides qualitative higher images.}
    \label{fig:real_life}
\end{figure}

Besides the scenes from RobustNeRF dataset, we test our method on the very common scenario, where a person acts as a distractor (\cref{fig:real_life}). The images are recorded with a smartphone and COLMAP~\cite{schoenberger2016sfm} is used to compute camera poses. 21\% of the training images contain a person. 
\\
While Gaussian Splatting suffers from artifacts caused by the distracting person, our method successfully removes the distractor. 

\end{document}